\pdfoutput=1
\documentclass[11pt]{article}
\usepackage[margin=1in]{geometry}
\usepackage{times}
\usepackage{amsmath,amssymb}
\usepackage{graphicx}
\usepackage{booktabs}
\usepackage{longtable}
\usepackage{tabularx}
\usepackage{subcaption}
\usepackage{enumitem}
\PassOptionsToPackage{hyphens}{url}
\usepackage[numbers,sort&compress]{natbib}

\usepackage{hyperref}
\usepackage{xcolor}
\usepackage[T1]{fontenc}
\usepackage{microtype}
\usepackage{xspace}
\usepackage{algorithm}
\usepackage{algpseudocode}
\usepackage{array}
\newcolumntype{R}[1]{>{\raggedright\arraybackslash}p{#1}}
\usepackage{float}
\usepackage{tikz}
\usetikzlibrary{positioning,arrows.meta,calc,fit,backgrounds}
\usepackage{pgfplots}
\pgfplotsset{compat=newest}
\definecolor{paceblue}{HTML}{2A78D6}
\definecolor{paceorange}{HTML}{EB6834}
\definecolor{paceink}{HTML}{1C1C1A}
\definecolor{pacemuted}{HTML}{6B6B66}
\definecolor{pacegrid}{HTML}{DCDCD6}
\tikzset{
  pacebox/.style={draw, rounded corners=2pt, line width=0.5pt, align=center,
                  inner sep=4pt, font=\scriptsize},
  paceflow/.style={-{Latex[length=1.6mm,width=1.2mm]}, line width=0.5pt},
}
\hypersetup{
  colorlinks=true,
  linkcolor=blue!50!black,
  citecolor=blue!50!black,
  urlcolor=blue!50!black
}

\newcommand{\pace}{\textsc{Pace}\xspace}
\newcommand{\scine}{\textsc{Scine}\xspace}

\title{\textbf{PACE: Precise AI Cinematic Expression} \\
\large A Typed Specification for Script-Grounded Previsualization and Geometric Conformance}

\author{
  Bing Duan\thanks{Corresponding author:
    \href{mailto:duanbing@pku.org.cn}{\texttt{duanbing@pku.org.cn}}} \quad
  Qiang Guo \quad Linpu Li \quad Zhijian Mao \quad Min Zhu \quad Zhirui Ren \\
  Yiwei Yan \quad Xi Chu \quad Xiaoding Li \\[4pt]
  Studio $\pi$
}
\date{}

\begin{document}
\maketitle

\begin{abstract}
Between a screenplay and a film made to an industrial standard sits a planning
problem that is spatial first: who stands where, what a camera sees from where
it stands, and how much of that space one lens holds. The breakdown that should
carry a director's intent into that plan is slow by hand and has no structured
place to put the intent, so an
image diffusion model asked for a shot in free text settles the plan by its own
defaults. We present \pace{} (Precise AI Cinematic
Expression), a typed representation for the plan: the
evidence from the screenplay, the characters, props and locations it needs,
where each subject stands, and what the camera does. A value is written once at
the level it belongs to -- script, scene, shot or panel -- and every level below
inherits it, so a lens declared for a scene does not have to be restated for
each of its panels. A compiler turns the inherited result into two things at
once: the words sent to the diffusion model, and a 3D scene built in metres. A camera
solver then works out where a camera must stand for the declared framing to be
the framing that is built.
Where a declared value becomes geometry rather than a word in a prompt,
what was built can be measured against what was asked for: \pace{} reports, field
by field, how far the compiled camera and the staged render sit from the
declaration, by measuring them rather than by asking a model to judge.

On the 11-scene \emph{Automatic Drive} screenplay, a scoring model rates the
breakdown at 0.95 coverage, against 79 events another model read out of the
screenplay. That is too high: when the audit deleted an action, the scorer
moved its event to the next one and scored it partly right instead of missing. Every staged panel with one
subject in it puts that subject within 1.2\% of frame width of the horizontal
position declared for it. With two or three subjects, one camera pose cannot
satisfy every declared position at once, and the error that is left over is
reported rather than absorbed. On
204 external director-storyboard shots, delivered head height is 1.906 times
the staged target from the director's own words, 1.733 from the compiled prompt,
and 0.955 with the greybox control. The condition that holds the declared framing
best is the one that draws the described action least, and vice versa. Declaring
the pose recovers part of that loss without giving the framing back: on the 30
shots whose pose the specification states, staging it takes the action drawn
from 58.9\% to 74.4\%, and the framing does not move.
The study also identifies current limits:
screen positions are derived from cast order, transitions are not implemented,
none of the 76 subject motion vectors are fitted to the action written for them, only set from coarse presets, and holistic human review of
the generated panels remains unrun. Code:
\url{https://github.com/StudioPiLabs/pace-core}.
\end{abstract}

\section{Introduction}
\label{sec:intro}

Between a screenplay and a film made to an industrial standard sits a planning problem that is spatial before it is anything else: who stands where, what a camera can see from where it stands, and how much of that space one lens holds. Two things make it hard to do with a generative model in the loop: the breakdown that should carry a director's intent into that plan is slow to produce by hand and has no structured place to put the intent, and an image diffusion model asked for a shot in free text settles the plan by its own defaults. Storyboarding and previsualization each own part of the plan, and closing the gap between them takes a system that is both a structured storyboard breakdown and an explicit 3D scene a camera can move through. A production storyboard breaks a script down into scene, characters, props, locations, and \emph{direction}: how each character is staged and blocked within the frame, shot size, camera movement, and how one shot transitions to the next. Storyboard software structures few of them: Toon Boom Storyboard Pro, the industry-standard tool, carries four caption fields in its default panel schema and no dedicated field for cast, props, location or shot size; camera movement is keyframed on a 2D frame, and a scene is 2D by default~\cite{toonboom_help}. Its one controlled vocabulary is transitions (Cut, Dissolve and three Wipe variants), the precedent for \pace{}'s proposed transitions field (Section~\ref{sec:schema}). Previsualization, in the ASC-ADG-VES definition,\footnote{The joint subcommittee of the American Society of Cinematographers (ASC), the Art Directors Guild (ADG), and the Visual Effects Society (VES), convened to standardize previsualization terminology; its definitions are the ones the Previsualization Society adopted~\cite{previs_ves}.} is the related discipline that plans camera placement, staging and editing ``predominantly using 3D animation tools and a virtual environment'', often starting from a screenplay with no storyboard passed to it at all~\cite{vfxvoice_previs}.

\textbf{Three consequences.} Three failures follow from leaving those decisions in free text, and this paper measures all three on its own corpus. First, a value stated in free text has nowhere to live, so it is restated for every shot that needs it, and what the restatements carry drifts from what is staged. One sentence is enough to matter: in scene~2 the location description names a console the delivered frame never shows, and across that scene's cut the same seeds move 53\% of their flat-share level with that sentence in the prompt against 18\% without it (Section~\ref{sec:look-continuity}). Nothing about the sentence is wrong as English, and no check over the text can catch it, because the thing it disagrees with is the scene, and the scene is not written down anywhere to be compared against. Second, a framing left to free text is not undecided, it is decided by the diffusion model: asked in the director's own words, the delivered head comes back at 1.906 times the height the same shot was staged at, and the declared order of shot sizes survives in 82.0\% of same-seed pairs against 97.1\% when the shot is staged first (Section~\ref{sec:prose-baseline}). Third, a panel described in free text has nothing to be measured against, so only a person looking at it can say whether it did what was asked. Figure~\ref{fig:one-scene} shows one scene of the corpus built the other way: its setups are staged once, and every panel inherits the same room and the same people.

\textbf{Three obstacles.} They stand between a screenplay and a
panel whose geometry can be checked. First, a screenplay under-determines
staging: it names who is present and what happens, but the screenplay evaluated
here contains no camera term, so a director or planner must supply shot size,
angle, and position as authored data. Facts that are extracted, rather than
authored, must remain traceable to the words that support them. Second, the
current solver deliberately keeps physical blocking, shot size, camera angle,
and roll fixed. Only pan and tilt remain, so one screen-space target is exactly
determined and several targets generally require a compromise. This is a
constraint induced by the present design, not a claim that multi-subject camera
placement is intrinsically infeasible; joint camera-and-staging methods retain
more degrees of freedom (Section~\ref{sec:related}). Third, a diffusion model may not
carry control faithfully: a state declared in text can fail to reach the image,
and the style of a render can follow the shading of its control image. Geometry
can therefore be checked before generation, while claims about delivered
appearance require measurements on the generated pixels.

\pace{} answers them with one mechanism: every decision a later stage must act on is a typed field from a fixed vocabulary carrying a unit, written once at the level it belongs to, inherited down script $\to$ scene $\to$ shot $\to$ panel, and resolved into an explicit metric 3D scene, so it can be solved for before the panel is rendered and measured on the pixels after. Eight previs-facing controls (scene, characters, props, locations, and the four that make up direction) drive that scene: character and prop proxies staged and physically simulated into rest poses, a camera trajectory compiled from the same representation, and a depth-driven control signal that a video-diffusion backbone renders from rather than infers. Throughout, \pace{} names the specification and the pipeline that fills, stages and checks it, and Studio $\pi$ the project it was built in. The premise is \scine{}'s human study (20{,}000 videos, 13 models), which finds that text-to-video models degrade substantially on Camera and especially Events~\cite{scine2025}. Camera work, blocking and continuity are therefore authored and simulated explicitly, and the generative model renders texture and motion over an already-specified geometry, as structural conditioning does in image diffusion~\cite{controlnet2023}. \pace{} has 146 authoring leaf fields in all; this paper reports on the previs-facing subset.

\textbf{Role of language models.} A large language model (LLM) performs the two
steps that do not reduce to parsing: decomposing the screenplay into per-scene
structure with character and prop registries (Section~\ref{sec:previs-recipe}),
and resolving free-form camera language into \pace{}'s motion domain-specific
language (DSL; Section~\ref{sec:geometry}). Compilation and staging after these
calls are deterministic and run in a fixed order
(Algorithm~\ref{alg:pace-pipeline}); generation and LLM-based evaluation are
stochastic and are reported separately.

\textbf{Scope.} This paper follows three implemented controls through the
artifact chain: scene boundaries (\emph{script breakdown}), actor blocking
(\emph{composition}), and shot size and lens (\emph{camera}). It also measures
continuity across existing cuts, but the transitions field itself is not
implemented (Section~\ref{sec:transition-adherence}). Geometric fields are
measured on compiled cameras and staged renders. Only the explicitly named
style, state, framing, and content experiments inspect generated pixels, and
holistic visual adherence remains unmeasured. Lighting and character attributes
beyond blocking are compiled into prompts but are not evaluated as general
controls.

Concretely, \pace{} makes three contributions:

\begin{enumerate}[leftmargin=*]
  \item \textbf{A typed previs representation and artifact contract}
  (Section~\ref{sec:pai}). Scene, characters, props, locations, and direction
  are inherited through script, scene, shot, and panel levels, then compiled
  into backend prompts and a metric scene.
  \item \textbf{Evidence-backed population and geometric compilation.} Extracted
  events and state changes retain quotations from the screenplay
  (Section~\ref{sec:verified-breakdown}); authored camera and blocking fields
  compile into a 3D scene. The camera solve itself is classical
  (Section~\ref{sec:related}); the contribution is the path from a typed,
  provenance-bearing document to a read-point target, a staged scene, and a
  field-specific residual (Section~\ref{sec:composition-precision}).
  \item \textbf{A layered conformance audit.} Typed and geometric fields are
  checked on the record, compiled camera, or staged render without a learned
  judge. Generated-panel measurements are kept separate and name their proxy or
  external instrument. Table~\ref{tab:results-summary} reports positive and
  negative results at the artifact level where each is observed
  (Section~\ref{sec:self-oracle}).
\end{enumerate}

\section{Related Work}
\label{sec:related}

Five bodies of work bear on the three failures of Section~\ref{sec:intro}. Each is placed by which failure it addresses and which it leaves open.

\textbf{Structure the discipline already specifies.} Screen-direction pedagogy treats coverage, staging and continuity as checkable structure: Robotham derives a scene's coverage from its tagged action, gaze, dialogue, location, props and movement~\cite{robotham2021}, Katz specifies blocking by distance, height, occlusion and grouping~\cite{katz2019}, and continuity is a rule set of the 180-degree line, the 30-degree rule and the eyeline match~\cite{bowen2023}. The Prose Storyboard Language gives shots a machine-readable vocabulary~\cite{psl2015}, and Louarn et al.\ stage characters and cameras jointly from spatial relations~\cite{louarn2018}. \pace{} keeps that decomposition as typed fields, filled from a screenplay with evidence, inherited across four levels and checked on the frame. Generation systems repair the same drift in the prompt layer, SEAM with a per-shot entity memory~\cite{seam2026} and CANVAS with continuity anchors~\cite{canvas2026}; \pace{} removes the restatement instead, writing a value once for every panel that inherits it. \pace{} hands the solver the numbers and the backend the resulting metric scene~\cite{vace2025}, so whether a number is obeyed does not arise, and the staged scene can go to a diffusion model, a character model or a crew.

\textbf{Turning a screenplay into structured data.} Screenplays have been parsed into typed elements~\cite{agarwal2014} and into per-scene JSON with character offsets~\cite{screenpy2017}, without checking the offsets; the parse here slices each range back out (Section~\ref{sec:breakdown-formal}). Verified quoting is established in GopherCite~\cite{gophercite2022}, TANL~\cite{tanl2021} and IGCS~\cite{igcs2025}, and a persistent state timeline goes back to the frame axiom and to threat detection in narrative planning~\cite{riedl2010}, with ProStruct~\cite{prostruct2018} and TRIP~\cite{trip2021} constraining state change in text. On screenplays, STAGE recovers each character's state across 151 screenplays~\cite{stage2026}, ATLAS compares world graphs to flag hallucination~\cite{atlas2026}, and Dramatron carries no entity state~\cite{dramatron2023}; here the same fields drive a renderer, so a dropped event is a missing panel. A storyboard-conditioned \textsc{Stage}~\cite{stagestoryboard2026} and ShotDirector~\cite{shotdirector2026} take the storyboard as given and control the transition in the video model, where \pace{} produces the storyboard. Cutting at a change of state follows Zehe et al.'s scene boundaries~\cite{zehe2021} at a finer grain than turning points~\cite{tripod2019}, read off extracted state rather than surface text, and not scored against a human reference (Section~\ref{sec:beats}). What lacks precedent is the measurement: MovieAgent~\cite{movieagent2025} and FilmAgent~\cite{filmagent2025} judge a breakdown through what it renders, SAGE~\cite{sage2026} scores its text with a rubric validated against directors, and none of the three asks whether a breakdown covers its screenplay. Section~\ref{sec:verified-breakdown} does, with planted errors against a control run~\cite{fbi2024,mutationjudge2026} and a judge whose agreement with humans is only approximate~\cite{mtbench2023}.

\textbf{Taxonomies, control and benchmarks.} \scine{}~\cite{scine2025} is an evaluation taxonomy of 76 leaves with a learned evaluator; \pace{} is an authoring schema (Section~\ref{sec:schema}). LaMP's camera DSL~\cite{lamp2026} is ported here as one compiler, resolved by an off-the-shelf model, beside a library of named moves with evidence grades (Section~\ref{sec:trajectory}); CameraCtrl~\cite{cameractrl2024} and viewpoint tokens~\cite{viewpointtokens2026} instead hand the camera to the model as a signal or an embedding. \pace{} leaves the backbone fixed~\cite{vace2025,wan} and contributes the simulated control signal. The closest camera check post-trains a video model against a geometry reward from another clip's trajectory~\cite{camerageomreward2025}; \pace{} measures a delivered frame against a declared angle, lens and size, on an untrained model. ShotBench~\cite{shotbench2025} and MovieNet~\cite{movienet} measure what a model reads or produces, not whether a declared value arrived.

\textbf{Sources of the camera prior.} One line learns the prior from a corpus in which the camera is annotated~\cite{cinesurvey2024,cameratrajsurvey2025}, as director-of-photography models~\cite{gendop2025} and shot-annotated pairs~\cite{cinetrans2025} do. The other solves the camera, and \pace{} belongs to it: screen-space constraints~\cite{gleicher1992}, idioms as placement programs~\cite{he1996}, a survey~\cite{christie2008}, the Toric Space for two or three subjects~\cite{lino2015} and a subject-aware model trained on simulation~\cite{lenscraft2025}. That tradition places a camera in a given scene; \pace{} also constructs the scene from a typed specification, aims at a chosen read point, and verifies on the rendered frame.

\textbf{Planning shots with agents.} VERTIGO optimizes camera previews against a learned judge~\cite{vertigo2026}, Rao et al.\ rank engine-rendered shots with a trained discriminator~\cite{virtualstoryboard2023}, and CinePreGen~\cite{cinepregen2024}, CineVision~\cite{cinevision2025} and PrevizWhiz~\cite{previzwhiz2026} reach previs interactively. Language-model planners work without a scene~\cite{videodirectorgpt2024,movieagent2025}. Systems that stage the plan in 3D differ in how they check it; StoryBlender~\cite{storyblender2026} and CineCrew~\cite{cinecrew2026} are the closest, and \pace{} claims not a 3D feedback loop but its combination of screenplay quotations, inherited typed fields, compilation to a staged artifact, and residuals named for the declaring field, without a head-to-head comparison. SAGE~\cite{sage2026} learns whether a directing decision is good; \pace{} asks whether a declared one was delivered, and its released storyboards, PROSE, are the external corpus of Section~\ref{sec:prose-baseline}.

\section{Problem Formulation and Terminology}
\label{sec:goodprevis}

\textbf{Problem statement.} The input to \pace{} is a screenplay and the
project's registries of characters, props, and locations. The output is a set of
storyboard panels and, optionally, a short clip per shot, each traceable to the
\pace{} fields it was built from. The objective is that every declared field is
discharged into staged geometry or prompt text and checked at the artifact level
where that discharge is observable. Section~\ref{sec:pai} separates authored
edits, compilation, and field-specific conformance residuals
(Eq.~\ref{eq:cvla}). The rest of this section gives the criteria, design
principles, and artifact vocabulary.

\begin{figure}[!tbp]
  \centering
  \includegraphics[width=\textwidth]{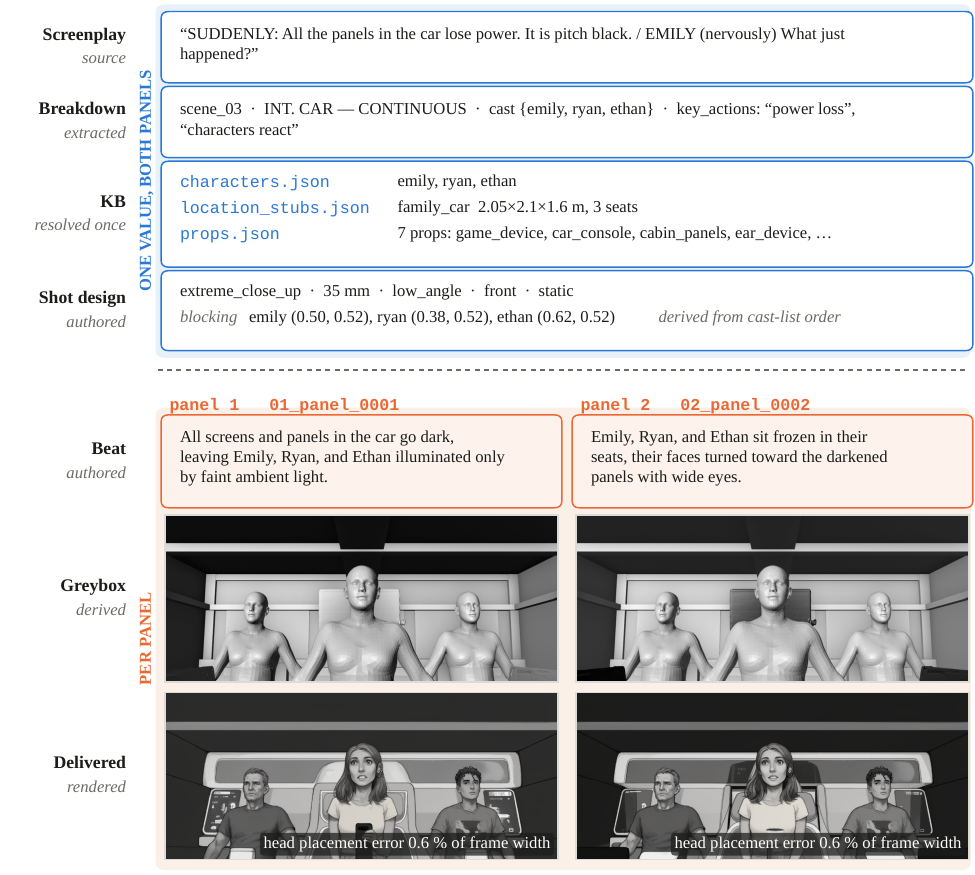}
  \caption{Consistency across a scene is structural: everything on the blue
  band is resolved once and read by both panels of \texttt{scene\_03}, so they
  cannot disagree about the cabin, the cast, the props or the lens, and only
  the beat on the orange band differs. Every value is read from the project's
  own files, each row names the mechanism that produced it, and the scene is
  shown before the composition step split its tight three-shots into singles
  (Section~\ref{sec:shot-size-adherence}). Three details are not offered as
  virtues. Emily's \emph{(nervously)} and the beat's \emph{wide eyes, frozen
  posture} reached no delivered face until the identity trait \emph{often
  smiling or laughing}, compiled into every panel, was removed. The blocking is identical in both
  panels because screen position is assigned by list order, a corpus-wide
  constant (Section~\ref{sec:field-provenance}), and the two greyboxes are
  pixel-identical because the scene declares one camera setup, a shotlist
  coverage gap (Section~\ref{sec:scene-to-panel}). Panel 1 is the beat in which the cabin loses power, so it declares the screens still on and stages them lit; panel 2 inherits them off, and the props it declares powered off are toned dark in its greybox, so the change reaches the image the sampler starts from and not only the prompt (Section~\ref{sec:visual-adherence}). Both renders share a seed and a denoise.}
  \label{fig:one-scene}
\end{figure}

Figure~\ref{fig:one-scene} follows one scene from screenplay to rendered panels and shows where its consistency comes from. It is keyed on \texttt{scene\_03\_shot\_01\_panel\_0001} (scene, shot, panel), before the composition step split the scene's tight three-shots into singles (Section~\ref{sec:shot-size-adherence}). Each row names the transformation of Section~\ref{sec:breakdown} that produced it. Measured as in Section~\ref{sec:composition-precision}, the panel's head placement error is 0.6\% of frame width. Section~\ref{sec:visual-adherence} returns to it to report where the chain breaks.

\label{sec:previs-criteria}
\textbf{Quality criteria.} Industry practice supplies the criteria for good previs. Beck's VES Handbook chapter names seven failure modes~\cite{veshandbook}, from which five criteria follow: a good previs is cheap and fast relative to the production, stays revisable, and asks only for moves that physics and rigs can deliver. It uses the production's real camera and lens data and serves several departments. Previs pays by moving hard-to-reverse decisions earlier, and generative production needs the same discipline. 

\label{sec:principles}
\textbf{Three principles the pipeline is evaluated against.} Three conventions of storyboard and directing pedagogy fix what \pace{} must carry, and stating them in advance makes the system checkable. P1 is discharged by the panel audit of Section~\ref{sec:breakdown-eval}, P2 by Sections~\ref{sec:pai} and~\ref{sec:geometry}, and P3 by Section~\ref{sec:composition}, measured on rendered output.

\textbf{P1. State change is the automatic segmenter's prior.} The pipeline
proposes a panel boundary when dramatic state changes through new information,
an action phase, a reaction, or a rhythmic accent, rather than after a fixed
duration~\cite{bowen2023}. This is a design prior, not a universal rule of
editing: a shot-size change can itself express emphasis or rhythm. Accordingly,
the automatic checker treats an isolated size change as insufficient evidence
for a new semantic beat, while the evaluation leaves boundary quality open
until a human reference segmentation is available. Impact and trigger are
represented as short phases, and reaction and aftermath as longer phases.

\textbf{P2. What a later stage must act on is a vector, and it resolves to geometry.} Every setup covering a scene must show the same room and the same person, and a description reinterpreted per angle yields a new scene each time. Anything that a downstream stage depends on is therefore a typed, unit-bearing value resolving to 3D geometry, reproducible as a plan view~\cite{katz2019}: where a character stands, which way a body faces, how far the camera sits and through what focal length.

\textbf{Grounding tiers.} P2 defines the two grounding tiers used throughout the paper. A field is at \textbf{tier~1} when it reaches the diffusion model as prompt text, so only a text encoder acts on it. It is at \textbf{tier~2} when it resolves into the staged metric scene and reaches the renderer as geometry, through the compiled camera and the depth and segmentation passes. Tiers say where a value is discharged, not how important it is: gaze is tier~1, rendered as an expression, while screen position is tier~2, because the camera aim is solved for it. A free-text descriptor sits below both.

\textbf{P3. One panel carries one primary read point, and the frame is accepted by a reading test.} The primary read point, the secondary read point and the environmental evidence are ranked, since spreading three statements evenly defocuses all of them. The frame passes a legibility test, not an aesthetic one: shrunk to stamp size, is the read point still findable~\cite{block2021}? The test concerns where the eye lands, not where the subject's bulk sits (Section~\ref{sec:composition-precision}).

\label{sec:artifact-vocab}
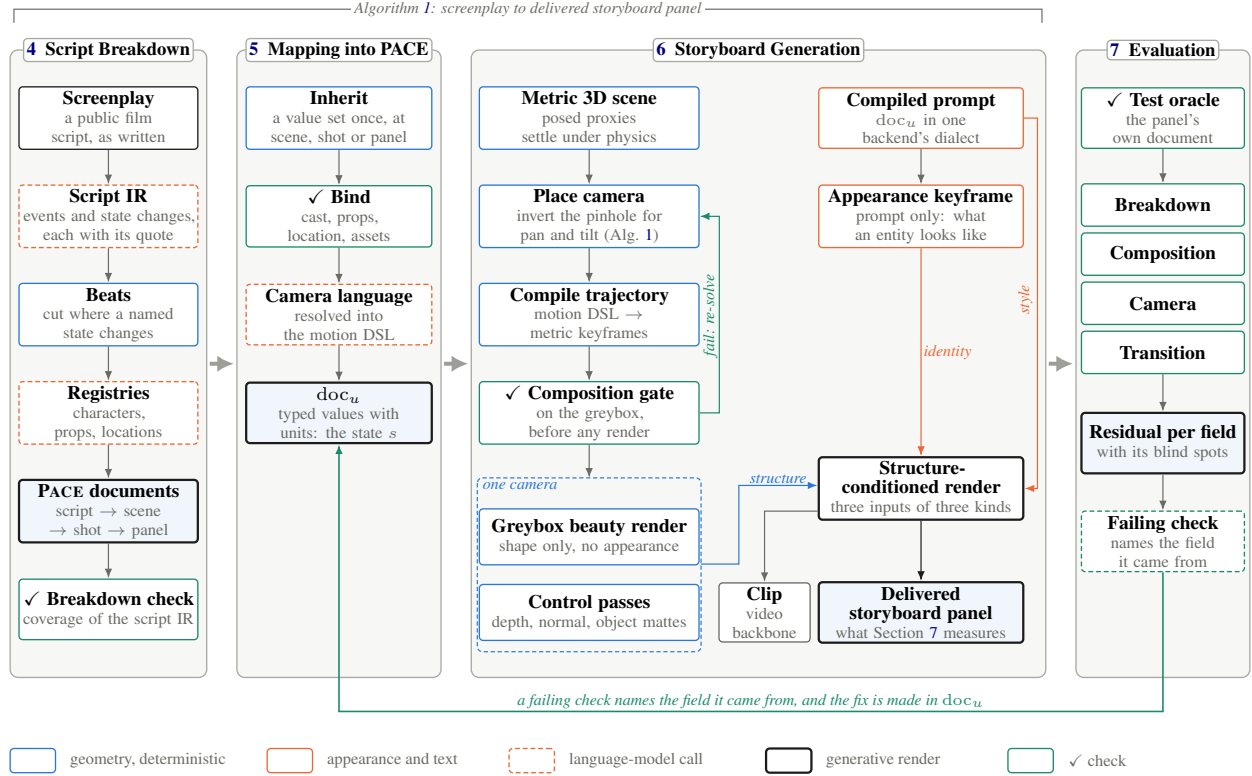
\begin{figure}[!tbp]
  \centering
  % The paper's overview: the four stages of Sections 3-6, in the order a panel
% passes through them, with both algorithms and the six named artifacts drawn
% inside the stage that produces them. TikZ like the other diagrams, so the
% figure is set in the document's own type and its section numbers are live
% \refs rather than transcribed.
%
% Encoding, carried by the legend and never by colour alone: geometry is blue
% and solid, appearance and text orange, a language-model call orange and
% dashed, a generative render ink and heavier, a check green and ticked.
% Hyphenation is off inside nodes: a 28 mm box broke "location" and "before"
% across lines, which reads worse than a slightly ragged line.
\definecolor{pacegreen}{HTML}{1A8A6A}
\newcommand{\ot}[1]{{\fontsize{7.2}{8.1}\selectfont\bfseries #1}}
\newcommand{\os}{\fontsize{6.2}{7.1}\selectfont\color{pacemuted}}
\begin{tikzpicture}[
  x=1mm, y=1mm,
  every node/.style={font=\fontsize{6.2}{7.1}\selectfont,
                     execute at begin node={\hyphenpenalty=10000\relax}},
  st/.style={draw=#1, rounded corners=1.4pt, line width=0.5pt, fill=white,
             align=center, inner sep=1.5pt, minimum height=8mm},
  geo/.style={st=paceblue},
  app/.style={st=paceorange},
  lm/.style={st=paceorange, dash pattern=on 1.8pt off 1.1pt},
  gen/.style={st=paceink, line width=0.85pt},
  chk/.style={st=pacegreen},
  res/.style={st=paceink, line width=0.85pt, fill=paceblue!7},
  sw/.style={minimum width=6mm, minimum height=3.2mm, inner sep=0pt},
  flow/.style={-{Latex[length=1.3mm,width=1mm]}, line width=0.45pt,
               draw=pacemuted},
  big/.style={-{Latex[length=2.2mm,width=2.2mm]}, line width=1.5pt,
              draw=pacegrid!45!pacemuted},
  stage/.style={draw=pacegrid!55!pacemuted, fill=pacegrid!22,
                rounded corners=3pt, line width=0.5pt},
  tab/.style={draw=pacegrid!55!pacemuted, fill=white, rounded corners=1.5pt,
              line width=0.5pt, inner sep=1.2pt,
              font=\fontsize{7}{8}\selectfont\bfseries},
  lab/.style={font=\fontsize{6}{7}\selectfont\itshape, text=pacemuted,
              inner sep=0.8pt},
]
  % ---- Algorithm 1 spans the first three stages; Algorithm 2 is one step of it
  \draw[pacemuted, line width=0.45pt] (0.5,0.6) -- (0.5,2.4) -- (136.5,2.4) -- (136.5,0.6);
  \node[lab, fill=white, inner sep=1.2pt] at (68.5,2.4)
    {Algorithm~\ref{alg:pace-pipeline}: screenplay to delivered storyboard panel};

  % ---- stage containers (equal height, so the flow between them is level)
  \begin{scope}[on background layer]
    \draw[stage] (0,-3)   rectangle (26,-86);
    \draw[stage] (30,-3)  rectangle (57,-86);
    \draw[stage] (61,-3)  rectangle (137,-86);
    \draw[stage] (141,-3) rectangle (164,-86);
  \end{scope}
  \node[tab] at (13,-3)    {\ref{sec:breakdown}\enspace Script Breakdown};
  \node[tab] at (43.5,-3)  {\ref{sec:pai}\enspace Mapping into PACE};
  \node[tab] at (99,-3)    {\ref{sec:geometry}\enspace Storyboard Generation};
  \node[tab] at (152.5,-3) {\ref{sec:evaluation}\enspace Evaluation};

  % ---- Section 3: the screenplay becomes documents
  \begin{scope}[every node/.append style={text width=22.6mm}]
    \node[st=paceink] (s0) at (13,-12)
      {\ot{Screenplay}\\ \os a public film script, as written};
    \node[lm]  (s1) at (13,-25)
      {\ot{Script IR}\\ \os events and state changes, each with its quote};
    \node[geo] (s2) at (13,-38)
      {\ot{Beats}\\ \os cut where a named state changes};
    \node[lm]  (s3) at (13,-51)
      {\ot{Registries}\\ \os characters, props, locations};
    \node[res] (s4) at (13,-64)
      {\ot{\pace{} documents}\\ \os script $\to$ scene $\to$ shot $\to$ panel};
    \node[chk] (s5) at (13,-77)
      {\ot{\checkmark\ Breakdown check}\\ \os coverage of the script IR};
  \end{scope}
  \foreach \a/\b in {s0/s1, s1/s2, s2/s3, s3/s4, s4/s5}
    \draw[flow] (\a) -- (\b);

  % ---- Section 4: one resolved document per panel
  \begin{scope}[every node/.append style={text width=23.5mm}]
    \node[geo] (p1) at (43.5,-12)
      {\ot{Inherit}\\ \os a value set once, at scene, shot or panel};
    \node[chk] (p2) at (43.5,-25)
      {\ot{\checkmark\ Bind}\\ \os cast, props, location, assets};
    \node[lm]  (p3) at (43.5,-38)
      {\ot{Camera language}\\ \os resolved into the motion DSL};
    \node[res] (p4) at (43.5,-51)
      {\ot{$\mathrm{doc}_u$}\\ \os typed values with units: the state $s$};
  \end{scope}
  \foreach \a/\b in {p1/p2, p2/p3, p3/p4}
    \draw[flow] (\a) -- (\b);

  % ---- Section 5, column A: geometry, solved and checked before a render
  \begin{scope}[every node/.append style={text width=28mm}]
    \node[geo] (a1) at (76.6,-12)
      {\ot{Metric 3D scene}\\ \os posed proxies settle under physics};
    \node[geo] (a2) at (76.6,-25)
      {\ot{Place camera}\\ \os invert the pinhole for pan and tilt (Alg.~\ref{alg:composition-solve})};
    \node[geo] (a3) at (76.6,-38)
      {\ot{Compile trajectory}\\ \os motion DSL $\to$ metric keyframes};
    \node[chk] (a4) at (76.6,-51)
      {\ot{\checkmark\ Composition gate}\\ \os on the greybox, before any render};
    \node[geo, minimum height=7.4mm] (a5) at (76.6,-67.4)
      {\ot{Greybox beauty render}\\ \os shape only, no appearance};
    \node[geo, minimum height=7.4mm] (a6) at (76.6,-77.4)
      {\ot{Control passes}\\ \os depth, normal, object mattes};
  \end{scope}
  \foreach \a/\b in {a1/a2, a2/a3, a3/a4}
    \draw[flow] (\a) -- (\b);
  % Both passes come off one camera, which is why the mattes register with the frame.
  \draw[draw=paceblue, dash pattern=on 1.6pt off 1pt, rounded corners=2pt, line width=0.45pt]
    (61.8,-59.6) rectangle (91.4,-82.6);
  \node[lab, text=paceblue, anchor=north west] at (62.2,-59.8) {one camera};
  \draw[flow] (a4.south) -- (76.6,-59.6);
  % The gate's failure path: back to the solve, never on to the renderer.
  \draw[flow, draw=pacegreen] (a4.east) -- ++(2.6,0) |- (a2.east);
  \node[lab, text=pacegreen, rotate=90, anchor=south] at (93.9,-38) {fail: re-solve};

  % ---- Section 5, column B: appearance, then the one render that joins them
  \begin{scope}[every node/.append style={text width=26mm}]
    \node[app] (b1) at (120.5,-12)
      {\ot{Compiled prompt}\\ \os $\mathrm{doc}_u$ in one backend's dialect};
    \node[app] (b2) at (120.5,-25)
      {\ot{Appearance keyframe}\\ \os prompt only: what an entity looks like};
    \node[gen] (b3) at (120.5,-61)
      {\ot{Structure-conditioned render}\\ \os three inputs of three kinds};
    \node[res] (b4) at (120.5,-77.4)
      {\ot{Delivered storyboard panel}\\ \os what Section~\ref{sec:evaluation} measures};
  \end{scope}
  \node[st=pacemuted, text width=11mm, minimum height=7.4mm] (clip) at (99.8,-77.4)
    {\ot{Clip}\\ \os video backbone};
  \draw[flow] (b1) -- (b2);
  \draw[flow, draw=paceorange] (b2) -- node[lab, right, text=paceorange] {identity} (b3);
  \draw[flow, draw=paceorange] (b1.east) -- (135.9,-12) |- (b3.east);
  \node[lab, text=paceorange, rotate=90, anchor=south] at (135.6,-36) {style};
  \draw[flow, draw=paceblue] (91.4,-71) -- (96.4,-71) |- ([yshift=0.4mm]b3.west);
  \node[lab, text=paceblue, anchor=south] at (101.6,-60.6) {structure};
  \draw[flow, draw=paceink] (b3) -- (b4);
  \draw[flow] ([yshift=-3mm]b3.west) -| (clip.north);

  % ---- Section 6: the document that asked for the panel checks it
  \begin{scope}[every node/.append style={text width=20.5mm}]
    \node[chk] (m1) at (152.5,-12)
      {\ot{\checkmark\ Test oracle}\\ \os the panel's own document};
    \node[res] (m3) at (152.5,-55)
      {\ot{Residual per field}\\ \os with its blind spots};
    \node[chk, dash pattern=on 1.8pt off 1.1pt] (m4) at (152.5,-68)
      {\ot{Failing check}\\ \os names the field it came from};
    \node[st=pacegreen, minimum height=5.6mm] (c1) at (152.5,-23.5) {\ot{Breakdown}};
    \node[st=pacegreen, minimum height=5.6mm] (c2) at (152.5,-30)   {\ot{Composition}};
    \node[st=pacegreen, minimum height=5.6mm] (c3) at (152.5,-36.5) {\ot{Camera}};
    \node[st=pacegreen, minimum height=5.6mm] (c4) at (152.5,-43)   {\ot{Transition}};
  \end{scope}
  \draw[flow] (m1) -- (c1);
  \draw[flow] (c4) -- (m3);
  \draw[flow] (m3) -- (m4);

  % ---- stage to stage
  \draw[big] (26.4,-44.5) -- (29.6,-44.5);
  \draw[big] (57.4,-44.5) -- (60.6,-44.5);
  \draw[big] (137.4,-44.5) -- (140.6,-44.5);

  % ---- the outer loop: a failed measurement is a fix to a field, not a reprompt
  \draw[flow, draw=pacegreen, line width=0.6pt] (m4.south) -- (152.5,-90.5) -| (p4.south);
  \node[lab, text=pacegreen, anchor=south] at (98,-90.4)
    {a failing check names the field it came from, and the fix is made in $\mathrm{doc}_u$};

  % ---- legend
  \node[geo, sw] at (3,-97) {};   \node[anchor=west, text=pacemuted] at (6.6,-97)   {geometry, deterministic};
  \node[app, sw] at (37,-97) {};  \node[anchor=west, text=pacemuted] at (40.6,-97)  {appearance and text};
  \node[lm, sw]  at (69,-97) {};  \node[anchor=west, text=pacemuted] at (72.6,-97)  {language-model call};
  \node[gen, sw] at (103,-97) {}; \node[anchor=west, text=pacemuted] at (106.6,-97) {generative render};
  \node[chk, sw] at (135,-97) {}; \node[anchor=west, text=pacemuted] at (138.6,-97) {\checkmark\ check};
\end{tikzpicture}
  \caption{\pace{} from screenplay to measured panel, with the six named
  artifacts in the order they arrive. The four stages are
  Sections~\ref{sec:breakdown}--\ref{sec:evaluation}:
  Algorithm~\ref{alg:pace-pipeline} runs from the screenplay to the delivered
  storyboard panel, and \textsc{PlaceCamera} in Algorithm~\ref{alg:pace-pipeline} is its
  camera-placement step. Two checks bracket the render. The composition gate
  reports a failing clause on the greybox, before anything is rendered, and
  its fix is a re-solve rather than a re-render; after the render, a
  measurement that fails names the field of $\mathrm{doc}_u$ it came from.
  The greybox beauty render and the control passes come off \emph{one}
  camera, which is why the mattes register with the frame rather than merely
  resembling it. The structure-conditioned render takes three inputs of three
  kinds (structure from the staged scene, identity from the appearance
  keyframe, style from the compiled prompt), and only the first is geometry,
  so only the first is measurable on the frame that comes back.}
  \label{fig:artifact-chain}
\end{figure}

\textbf{Artifact vocabulary.} ``The panel'' can mean a flat unstyled render of a staged scene or a fully styled still, so six named artifacts carry the recipe of Section~\ref{sec:previs-recipe} to a delivered image (Figure~\ref{fig:artifact-chain}). The \textbf{appearance keyframe} is prompt-only and fixes what an entity looks like. The \textbf{greybox beauty render} is a flat, unlit pass over the staged scene from the compiled camera, carrying subject count, pose and composition but no appearance. The \textbf{control passes} (depth, surface normal, per-subject mattes, optionally a face matte) come off the same camera. The \textbf{structure-conditioned render} starts from the greybox and control passes, with identity from the keyframe and style from the compiled prompt, and outputs the \textbf{delivered storyboard panel} (\texttt{panel\_$u$}, Algorithm~\ref{alg:pace-pipeline}) that Section~\ref{sec:composition-precision} measures. An optional pass on the same control signal yields a \textbf{clip}, a distinct artifact rather than a higher-frame-rate panel.

\section{Script Breakdown}
\label{sec:breakdown}

This section states how a screenplay becomes the object of Section~\ref{sec:pai}. It says which stage consists of model calls over the screenplay text, which is a deterministic scaffold, which is a per-scene model call against a closed vocabulary, and which is a checker that only reports.

Three transformations populate \pace{}, and a field's authority does not follow from its place in the schema. A field is \textbf{extracted} when a model call reads it off screenplay evidence (a scene heading, the cast present). It is \textbf{authored} when a person, or a model standing in for one, supplies a cinematic decision the screenplay leaves open (a shot size, a camera angle). It is \textbf{derived} when computed deterministically from other \pace{} fields (a solved camera rotation, a projected screen coordinate). Table~\ref{tab:provenance} gives each measured field's transformation. A 100\% fill rate on an authored field says a director did their job; on an extracted field it says the screenplay held the information.

\begin{table}[!tbp]
\centering
\footnotesize
\caption{Provenance of the fields this paper measures. \emph{Extracted} fields are read from screenplay evidence by a model call; \emph{authored} fields are a cinematic decision supplied by a person or a model standing in for one, because the screenplay does not settle them; \emph{derived} fields are computed deterministically from other \pace{} fields, with no further judgment. The same surface action, ``screenplay decomposition'', produces both extracted and authored fields in one call (Section~\ref{sec:scene-to-panel}); listing them together as one thing is the error Section~\ref{sec:breakdown} corrects. Screen position is derived rather than authored, which Section~\ref{sec:field-provenance} states and Section~\ref{sec:composition} gives the coordinates for.}
\label{tab:provenance}
\begin{tabularx}{\textwidth}{@{}R{2.9cm}R{3.7cm}R{1.9cm}X@{}}
\toprule
Source & \pace{} field & Authority & Downstream use \\
\midrule
scene heading & \texttt{backdrop.\allowbreak setting}, \texttt{.time\_of\_day} & Extracted & Prompt assembly, lighting defaults (Section~\ref{sec:scene-segmentation}) \\
cast in a scene & \texttt{subjects[].\allowbreak character\_id} & Extracted & Cast binding, reference conditioning \\
a listed key action & \texttt{actions[].\allowbreak description} (initial) & Extracted (label) & Panel count and boundary (Section~\ref{sec:scene-to-panel}) \\
that beat, elaborated & \texttt{actions[].\allowbreak description} (final) & Authored & Panel staging, body-seed action (Section~\ref{sec:enrichment}) \\
shot design & \texttt{shot\_size}, \texttt{angle}, \texttt{trajectory} & Authored & Unless the script names a camera term; lens, distance, camera pose (Section~\ref{sec:scene-to-panel}) \\
blocking (cast-list order) & \texttt{subjects[].\allowbreak screen\_position} zone & Derived & Composition-solver target (Section~\ref{sec:composition}) \\
solved camera rotation & compiled camera pose & Derived & Greybox render, depth/matte passes \\
screen-position zone & normalized $(x,y)$ target & Derived & $x \in \{0.38, 0.50, 0.62\}$, $y = 0.52$; input to the composition solve (\textsc{PlaceCamera}, Algorithm~\ref{alg:pace-pipeline}) \\
an entity's name & registry identifier & Extracted & Assigned once, then fixed; every downstream reference resolves against it (Section~\ref{sec:entity-identity}) \\
age-state likeness & \texttt{character@age\_state} & Authored & An editor assigns the plate; reference-conditioned render \\
\bottomrule
\end{tabularx}
\end{table}

\subsection{The breakdown as a formal object}
\label{sec:breakdown-formal}

Stating what a breakdown \emph{is} makes \emph{missing} and \emph{invented} set operations rather than opinions, which is what Section~\ref{sec:verified-breakdown} measures. Every definition below is the one \texttt{pace-core} implements: its breakdown verifier computes the coverage and fabrication counts of Eq.~\ref{eq:coverage}, and its beat segmenter the segmentation of Eq.~\ref{eq:cut}.

\textbf{Parse.} Write $\Sigma$ for the screenplay, as in Algorithm~\ref{alg:pace-pipeline}. A deterministic parse gives an ordered sequence of typed elements,
\begin{equation}
E(\Sigma) = \big\langle (\kappa_1, [a_1, b_1)), \ldots, (\kappa_n, [a_n, b_n)) \big\rangle,
\qquad \kappa_i \in \mathcal{E},
\label{eq:parse}
\end{equation}
Equation~\ref{eq:parse} preserves both order and source ranges.
Here $\mathcal{E}$ is the screenplay's own element vocabulary, \{scene heading, action, character cue, parenthetical, dialogue, transition, super\}, and each range is verified by slicing it back out of $\Sigma$ and comparing. This corpus gives $n = 242$. The round trip checks the ranges, not the labels: an element given the wrong $\kappa$ slices back out exactly as an element given the right one does. Tagging accuracy against a reference parse is not measured here, which is the half of Agarwal et al.'s evaluation~\cite{agarwal2014} this does not reproduce.

\textbf{Grounding.} A model reads $E(\Sigma)$ into a script IR, the triple $(N, V, X)$ of entities, events and the state changes the events carry, and is never asked for an offset, since a language model counts characters poorly. Each fact $z$ carries a quote $q(z)$, and grounding locates it by string search inside the span of the scene $z$ is claimed in, once verbatim and once with whitespace collapsed:
\begin{equation}
\mathrm{grounded}(z) \iff q(z) \sqsubseteq \Sigma\big[\mathrm{scene}(z)\big],
\label{eq:grounded}
\end{equation}
writing $\sqsubseteq$ for that search. A fact that fails is marked ungrounded and listed as a claim with no support, rather than repaired or silently kept: a quote found only in a different scene is not support for a fact in this one. On this corpus $|N| = 56$, $|V| = 79$ and $|X| = 22$, and the ungrounded list is empty.

What that check establishes is narrower than it looks, in two directions. It decides whether the quote is \emph{present}, not whether it \emph{supports} the fact attached to it: a model that quotes a real sentence and hangs a wrong fact on it passes. And the match is verbatim, so a correct fact whose quote was paraphrased is marked ungrounded, which is why IGCS relaxes the same test to a bounded edit distance under current models~\cite{igcs2025}. An empty list is therefore the weaker of the two readings available: on this corpus the check never fired, so it has shown that nothing was fabricated outright and has not yet shown that it would catch a fact whose quote is real and whose claim is not.

\textbf{World state.} Each $x \in X$ is a transition $(\text{entity}, \text{attribute}, \text{from}, \text{to})$ carrying its evidence and the location-continuity group it applies within. Applying them in script order gives $\omega_k$, what is true of every entity after the $k$th event. Two things it deliberately does not do: it does not merge entities across locations, since one vehicle's panels are not another's, and it does not guess when a state ends: a state persists until an event changes it, because that is what the screenplay's silence means. A beat then reads $\omega$ from this timeline rather than from its own events, which is the property that stops a later beat quietly restoring what an earlier one changed.

\textbf{Segmentation.} A beat segmentation partitions the ordered events into contiguous blocks. Each block $\beta_j$ is a triple $(\omega_{\text{before}}, \delta_j, \omega_{\text{after}})$: the state it starts in, its transition and the state it leaves; the partition is by construction, since the cut walks the events in script order and closes a block at each split. Six boundary features $f_g \in [0,1]$ are defined over an adjacent pair. Three are computed from the IR (\emph{state\_change}, whether the later event carries one; \emph{focus\_shift}, whether the actor differs; \emph{importance\_delta}), two are judged by an adjudicator model (\emph{goal\_shift}, \emph{reveal}), and one, \emph{spatial\_shift}, has no source in this IR, since events carry no location of their own, and is left unmeasured rather than invented from text distance. The score renormalizes the published weights over what was measured, so a corpus missing a feature does not quietly become unable to split:
\begin{equation}
S(f) = \frac{\sum_{g \in M} w_g\, f_g}{\sum_{g \in M} w_g},
\qquad
\mathrm{cut}(k) =
\begin{cases}
\textsc{split} & \exists\, g:\ f_g \ge \theta_g, \ \text{or } S(f) \ge \tau_{\mathrm{hi}},\\
\textsc{merge} & S(f) \le \tau_{\mathrm{lo}},\\
\textsc{adjudicate} & \text{otherwise,}
\end{cases}
\label{eq:cut}
\end{equation}
In Eq.~\ref{eq:cut}, $M$ is the set of measured features and $\theta = \{\text{state\_change}: 1.0,\ \text{goal\_shift}: 0.7,\ \text{reveal}: 0.7\}$ the signals sufficient on their own, and $\tau_{\mathrm{lo}} = 0.20$, $\tau_{\mathrm{hi}} = 0.55$. Renormalizing cuts both ways, and the second way is worth stating: dividing by the measured weight raises every score, so the thresholds are not the ones the published weights would meet. On this IR, where only the three structural features are measurable, the denominator is $0.60$ and every score is $1.67\times$ what the published weights give it. Two decisions turn on that factor. A lone \emph{focus\_shift} scores $0.33$ and is adjudicated where the published sum would merge it at $0.20$; \emph{state\_change} with \emph{focus\_shift} scores $0.83$ and splits where the published sum would adjudicate at $0.50$. The renormalization is therefore a choice to split more readily on a partly measurable corpus, not only a guard against splitting less. A merge and an unresolved adjudication both keep the events together, because over-splitting produces beats with no state change in them and those have nothing to depict. The first clause of $\mathrm{cut}$ is a measured choice, not a preference: under the weighted sum alone, 40 of 41 adjudicated pairs merged, and admitting a single strong signal took this film from 26 beats to 39 (Section~\ref{sec:beats}).

\textbf{Breakdown and its fidelity.} A breakdown is a map $D$ from beats to panels that fills the fields of Section~\ref{sec:pai}: cast from the block's participants, prop and light states from $\omega_{\text{after}}$, action text from its predicates, with the shot design authored beside them (Table~\ref{tab:provenance}). An estimator aligns each script event to $D$ with one of four labels, and the program, never the model, computes the rates. Writing $c(v, D) = 1$ for \textsc{exact} or \textsc{semantic}, $0.6$ for \textsc{over\_generalized} and $0$ for \textsc{missing}, the $0.6$ being a starting value rather than a measured one, so a coverage figure carries whatever that choice is worth,
\begin{equation}
\mathrm{cov}(D) = \frac{1}{|V|} \sum_{v \in V} c(v, D),
\qquad
\mathrm{fab}(D) = \big|\{\, d \in D : d \text{ asserts an event no } v \in V \text{ supports} \,\}\big|.
\label{eq:coverage}
\end{equation}
Equation~\ref{eq:coverage} keeps coverage and fabrication separate. An
over-specification, an entry that adds a framing the screenplay does not state, is counted separately and is not an error: a breakdown that writes ``a close-up of'' is doing its job.

Three properties follow, and they are what Section~\ref{sec:verified-breakdown} reports: \emph{partition}, every event in exactly one beat, which holds by construction and came out 79 of 79; \emph{persistence}, every state transition surviving segmentation, 22 of 22; and \emph{grounding}, Eq.~\ref{eq:grounded} for every fact the IR supplies. A breakdown written in one pass admits none of the three, because it produces no $(N, V, X)$ to hold itself to; the experiment measures that difference rather than a difference of wording.

\subsection{The stages that produce it}
\label{sec:breakdown-stages}

\label{sec:scene-segmentation}
\textbf{Scene segmentation.} Model calls return one scene per continuous unit of location, time of day and cast, the test a script supervisor uses, given to the model as a definition. Each scene carries a heading, cast, summary, key physical actions, and era, region and culture inferred from its own content; a field a call cannot support is returned null rather than guessed.

\label{sec:scene-to-panel}
\textbf{From scenes to shots and panels.} Expansion is deterministic scaffolding: each scene's \texttt{key\_actions} list becomes one shot and one starter panel per action, so the initial granularity is exactly that list's, and only whether the resulting panels differ is checked (Section~\ref{sec:breakdown-eval}). Shot size, angle and movement are pattern-matched from the action text, defaulting to medium, eye level, static. The pattern table was Chinese-only for most of the project, so no English action line matched it and every one fell to that default. The matcher therefore contributed no shot size to this corpus: the variety it has comes entirely from a director's shot design added afterward, which is why Table~\ref{tab:provenance} records shot size as authored rather than extracted. Transitions are not yet inferred: the corpus has no transition field (Section~\ref{sec:schema}), and only 4 of its 34 within-scene cuts change a declared camera value (Section~\ref{sec:transition-adherence}).

\label{sec:entity-identity}
\textbf{Character, prop, and location identity.} Separate model calls per entity class propose an identifier, a physical anchor description and class-specific fields. They merge into a project-wide registry that fills blanks but never overwrites a populated field, so a re-run cannot rewrite a human edit. Later shots resolve against the identifier, not the screenplay's spelling, and age state rides on it (\texttt{emily@adult\_50}). A location string that fails exact, substring and token-overlap matching stays unresolved rather than coerced. Garments are extracted as props, not as prose on a character: a production sources and hands over a costume the way it does a lantern, so it takes an identifier and a set of continuity states at the same moment every other prop does, and a shot names it with \texttt{costume\_\allowbreak id} rather than restating it. A description written once per shot is not comparable with the description in the next shot, which is what a cut needs (Section~\ref{sec:greybox-gate}).

\label{sec:enrichment}
\textbf{Enrichment against a closed vocabulary.} The scaffold leaves most leaf fields null, since neither the scene-segmentation calls nor the deterministic expansion populates them. A model call per scene fills them, with its output vocabulary taken from the schema's own enum definitions so the two cannot drift. An out-of-vocabulary token is dropped and the field left null, so a wrong answer shows as an absence.

\label{sec:beat-gates}
\textbf{Quality gates on the breakdown itself.} Following P1 (Section~\ref{sec:principles}), the audit in Section~\ref{sec:breakdown-eval} checks that no two adjacent panels agree on every authored field. It cannot see a beat that gives a director nothing to stage, packs two actions into one panel, or has ambiguous pronouns. A separate checker reads each beat against four rules, each named for a failure this corpus produced before the rule existed; it reports rather than rewrites, because what a label should say is a directorial decision. On the current corpus it flags nothing (Section~\ref{sec:evaluation}).

\section{Mapping into PACE}
\label{sec:pai}

Section~\ref{sec:formal} states the object formally,
Section~\ref{sec:schema} instantiates the state as a schema, and
Section~\ref{sec:holds-delivers} states what the declaration guarantees and
where delivery fell short.

\subsection{Typed state, compilation, and conformance residuals}
\label{sec:formal}

The formalism separates four operations that the implementation keeps
separate: editing a declaration, solving the free camera variables, compiling
render inputs, and measuring what the compiler produced. A \emph{document} is
a \pace{} record at the script, scene, shot or panel level, whose fields the
levels beneath it inherit; a \emph{registry}, one per entity class, assigns an
identifier once and never overwrites a populated field. Write $s$ for one
resolved panel record, $\Delta$ for an authored edit, and
$q=(\mathrm{yaw},\mathrm{pitch},d)$ for the camera variables left free after
the declared position, lens, and roll are resolved: two rotations and a lateral
offset $d$ along the camera's own right axis. Let $U$ apply an edit, $H$
implement the primary-read-point solve, $A$ the arbitration that follows it,
$F$ compile a resolved record at a specified $q$, and $g_j$ measure one named
contract on the resulting artifacts $a$:
\begin{equation}
\begin{gathered}
s' = U(s,\Delta), \qquad q^0 = H(s'), \qquad q^* = A(s',q^0), \\[2pt]
a = F(s',q^*), \qquad r_j = g_j(s',a), \quad j=1,\ldots,J .
\end{gathered}
\label{eq:cvla}
\end{equation}
Each $r_j$ is reported with its unit and threshold and may be
\emph{unmeasurable}; it is not folded into a single quality score. For binding
geometric contracts, let $\Omega(s')$ be the allowed pan--tilt domain and define
the diagnostic feasible set
$\mathcal{Q}(s')=\{q\in\Omega(s'):g_j(s',F(s',q))\leq\varepsilon_j\ \forall j\}$.
The set may be empty. The current implementation does not search $\mathcal{Q}(s')$, and it does not
have to, because the two stages divide the problem. $H$ solves the primary read
point in $(\mathrm{yaw},\mathrm{pitch})$ and reports its residual and
convergence; on this corpus it converges on all 42 staged panels, to a residual
below $10^{-5}$ of frame width. $A$ then decides what a single frame owes the
subjects $H$ did not aim at. Writing $\hat{x}_i$ for subject $i$'s declared
screen $x$ and $x_i(d)$ for where the staged body projects at lateral offset
$d$,
\begin{equation}
A(s',q^0) = \operatorname*{arg\,min}_{|d| \leq D}\;
  (1-w)\,\frac{1}{n}\sum_{i=1}^{n}\bigl|x_i(d)-\hat{x}_i\bigr|
  \;+\; w\,\max_{i}\bigl|x_i(d)-\hat{x}_i\bigr| ,
\label{eq:arbitration}
\end{equation}
searched on a grid ($D = 1.6$\,m at $1$\,cm, 321 probes). The offset is a
lateral dolly rather than a pan: the camera slides along its own right axis and
keeps its direction, so the fit distance, and with it the declared shot size,
is unchanged and the trade is paid in placement alone. The weight $w$ is the
arbitration this system would otherwise leave implicit, and it is a declared
constant at $0.5$ rather than a rule inside the solve. At $w=0$ the objective
is the plain sum, which cannot distinguish everyone slightly off from one
subject exact and one ruined, and reliably chooses the second: it reproduces a
focus-centred aim and leaves 7 of this corpus's 76 staged subjects outside the
frame. Raising $w$ to $0.30$ costs $0.70$ points of mean placement error, keeps
all 7 subjects in frame, takes the worst-served subject from $13.9\%$ to
$10.0\%$ and the best-to-worst gap from $13.8\%$ to $4.9\%$; declared
left-to-right order holds at 29 of 29 throughout and delivered body height moves
by under $0.2\%$. The frontier is flat above $0.30$. This is why a
multi-subject residual is shared rather than parked on one subject, and it is a
choice a production should be able to set per panel rather than a property of
the geometry. An empty joint feasible set still yields a compiled artifact
together with field-specific residuals, never an approximate result relabelled
as feasible.
Locked fields constrain $U$, while the grounding tiers state which fields reach
$F$ as geometry and which reach it only as prompt text.

The camera solve illustrates the distinction. Under the present design, physical
blocking, lens, and roll are fixed before \textsc{PlaceCamera}; pan, tilt and
the lateral offset are the free variables. One screen-space target therefore
gives two scalar constraints and can be solved to tolerance. Several subject
targets give more scalar constraints than one frame can satisfy, and the two
stages answer that in different ways: $H$ solves the target P3 selects and
reports its residual, and $A$ then trades that exact aim for a lateral offset
minimizing Eq.~\ref{eq:arbitration} over every declared position. Which subject
gives way is therefore a stated weight rather than a consequence of which
subject the read point named. This behavior does not establish that multi-subject camera
placement is intrinsically infeasible. Methods that jointly vary camera
position, lens, or blocking operate over a larger feasible region
(Section~\ref{sec:related}).

Parts of $F$ are computable from known geometry. Given camera pose, lens, and
subject positions, projection and continuity relations are derived rather than
predicted (Section~\ref{sec:composition}). Intent and a director's departures
from convention are not derivable from those quantities and remain explicit in
the registries and locks. Algorithm~\ref{alg:pace-pipeline} implements $H$ and
$F$ for the evaluated pipeline; $U$ is the authoring operation that precedes
them. The paper reports compilation and conformance, not a learned policy or
optimizer for choosing a cinematically preferable camera from
$\mathcal{Q}(s')$.

The notation is most useful on the negative results, because it separates four
kinds of failure that take different repairs. \textbf{A well-formed $s$ can carry
no information:} screen position is assigned by list order through a zone table
rather than authored (Section~\ref{sec:field-provenance}), so a score against it
measures self-consistency rather than fidelity. \textbf{$F$ can fail invisibly
to any audit of $s$:} some fields are in $s$ and never reach $a$. \textbf{A failure can lie outside the geometric contracts by construction:}
no predicate reaches image content, so
neither the style divergence (Section~\ref{sec:visual-adherence}) nor the seat
belt could have been caught.
\textbf{The instrument can fail:} the breakdown estimator re-attaches a deleted
event rather than reporting it dropped (Section~\ref{sec:verified-breakdown}),
which bounds every number it produced. A single quality score would have
averaged these four together.

\textbf{Predictions at the current operating point.} The separation above
makes three testable predictions. First, the two-angle solve should drive one
screen-space point to tolerance, while additional declared points may retain
residual under fixed staging. Section~\ref{sec:placement-budget} measures a
single subject at 0.3\% of frame width on staged geometry, averaged over 19
placements, and multi-subject panels at an order of magnitude more; the
degree-of-freedom count does not predict the difference between two- and
three-subject cases, 3.2\% versus 5.4\%. Second, a predicate defined only on the
compiled camera and staged geometry cannot establish conformance of generated
image content. Third, the geometric part of $F$ needs no learned camera prior:
it resolves composition on an external screenplay's panels from explicit scene
quantities.

\subsection{\pace{}'s pillars relative to \scine{}}
\label{sec:schema}

\pace{} is an authoring schema whose documents serialize into four containers
matching \scine{}'s pillars; the authoring view groups the same fields into
four pillars of 146 leaf fields in fully expanded form: \textbf{Characters} (42),
\textbf{Setup} (58), \textbf{Camera \& Lighting} (42) and \textbf{Events} (4). Every field count in this paper is of leaves of that authoring view; the serialization declares more typed fields than this, because it counts the containers they hang from as well. Fields are gathered by \emph{who has
to stay consistent across shots}: one person recurs in dozens of shots separated
by weeks, while a camera's settings are decided again in each shot.
The evaluation taxonomy \scine{}~\cite{scine2025} keeps per-subject attributes
in Setup and per-action, per-emotion and per-dialogue-line fields in Events,
where \pace{} gathers them under Characters, and separates Camera from Lighting,
where \pace{} merges them; comparisons to its published numbers use its
boundaries.

Of the eight core controls named in Section~\ref{sec:intro}, scene, locations
and props map onto Setup; characters and actor blocking onto Characters'
Identity \& Staging sub-group; and shot size and camera movement onto Camera \&
Lighting. The eighth, \emph{shot transitions}, has no \scine{} home and is not
yet implemented: we propose it as a controlled-vocabulary field, with
Storyboard Pro's Cut/Dissolve/Wipe vocabulary as the precedent (Section~\ref{sec:related}). Lighting, and the character fields beyond
blocking, are counted in the schema but outside this paper's scope
(Section~\ref{sec:conclusion}). The document hierarchy mirrors production
breakdown, \textbf{Script} $\to$ \textbf{Scene} $\to$ \textbf{Shot} $\to$
\textbf{Panel}, with inheritance resolved as shot defaults $\to$ shot $\to$
panel overrides $\to$ per-panel compile hints.

\subsection{What the specification holds, and what it failed to deliver}
\label{sec:holds-delivers}

The schema fixes what a panel declares; this subsection states what that declaration guarantees, and where the delivery fell short of it.

\textbf{Scope.} A consistency that is constructed is a guarantee, and one that
is only measured is not. Body shape is constructed: each character's SMPL-X
shape coefficients follow from the age state and build it declares, so the same
vector appears wherever that character does. Locations and props are
constructed as geometry but sampled as appearance. The control is also
one-directional: a field can require presence but not absence, so an unnamed
seat belt appears in one panel of a scene and not its neighbors. A typed
schema over a generative renderer governs what must be present, not what must be
absent.

\textbf{Two ways of holding a scene.} Rendering a scene's first panel once and
passing it to every later panel on the reference channel raised within-scene
layout agreement from $0.299$ to $0.524$ on one scene of four panels; the
method gives up rendering a scene's panels in parallel. Copying segmented characters between
panels makes identity pixel-identical but carries each panel's expressions and
poses into the next: identity is guaranteed and performance is lost.

\label{sec:resolved-not-used}
\textbf{Resolution is not use.} Field adherence and asset-anchor adherence both
scored 27/27 shots, correctly, yet eight resolved values did not reach the
diffusion model: four failed translation at the text-encoder boundary, three had no
path from the registry to the compiler, and in the eighth, screen position
reached the geometry but never the prompt, so on the corpus's three-subject panel the characters changed seats
while the positions held. Compiling the cast in screen order put all three in
their declared seats and reduced horizontal placement error from 2.8\% to
1.7\% of frame width. Every panel involved passed every referential check, which
is why this paper reports failures at the stage that produced them.

\section{Storyboard Generation}
\label{sec:geometry}

This section compiles a resolved \pace{} panel into the passes a renderer consumes. Section~\ref{sec:previs-recipe} traces one panel end to end; Sections~\ref{sec:composition} to \ref{sec:solver} fix the camera aim from the screen-position field; Section~\ref{sec:trajectory} extends the aim to a camera move; and Section~\ref{sec:events-smplx} grounds the event fields in a body model.

\subsection{Producing one panel, end to end}
\label{sec:previs-recipe}

The mechanisms below read more easily in the order they run. That order is not arbitrary: each step makes a decision that the next step cannot revise, which is the discipline professional practice imposes.

The breakdown finishes first, as its physical counterpart does. A character's registry entry fixes the trigger, anchor and per-age-state likeness that make that character reproducible in any shot covering them, so anything generated before that entry exists cannot be matched to anything generated after. Scenes are then cut into shots and panels, a panel being a pose or composition change within a shot rather than an elapsed interval.

The keyframe fixes appearance and only appearance; a change of setting, costume or light must appear there because nothing downstream can recover it. The 3D engine is specified as a capability set rather than a product: a scriptable scene graph and camera, rigid-body physics, and multi-pass rendering from an arbitrary camera path.

The same resolved panel document compiles into a body seed for every subject the panel asks to animate. Here, typed fields are parsed into geometry rather than handed to the renderer as text. \pace{} does not ask a renderer to infer a body from a sentence such as ``Ethan recoils''. It resolves the subject, age state, action, emotion, dialogue state, screen position, facing and gaze, shot duration and scene scale, then writes a structured seed in the SMPL-X/Motion-X convention~\cite{motionx2023} (Section~\ref{sec:events-smplx}). The seed is deliberately not final acting: a later fitting pass may replace its coarse channels, but must preserve the subject id, coordinate frame, panel read point and camera relation that \pace{} already approved.

One pass over the finished scene renders the greybox beauty pass, together
with the depth and segmentation passes that make the staged artifact checkable;
the structure-conditioned render then produces the delivered panel. Nothing in
this path recovers 3D from footage. Finally, geometric placement is measured on
the staged segmentation pass (Section~\ref{sec:composition-precision}).
Generated pixels are evaluated only in the explicitly named state, style,
framing, and content experiments of Section~\ref{sec:visual-adherence}; holistic
visual adherence is not measured.

Algorithm~\ref{alg:pace-pipeline} states this sequence as one procedure. It gives the computational dependency path rather than the deployed workflow; the human review gate between phases is real but omitted here.

\begin{algorithm}[tbp]
\caption{Compilation from a \pace{} document to previs artifacts.
The loop implements $H$ and $F$ in Eq.~\ref{eq:cvla}.
\textsc{PlaceCamera} implements $H$: it solves the primary read-point target
with pan and tilt and reports convergence; other declared subjects are
evaluated as residuals.}
\label{alg:pace-pipeline}
\label{alg:composition-solve}
\begin{algorithmic}[1]
\footnotesize
\Require Screenplay $\Sigma$; \pace{} schema version $\mathcal{S}$; character, prop, and location registries $\mathcal{R}_c, \mathcal{R}_p, \mathcal{R}_\ell$; backend compilers $\mathcal{K}$; geometry-conditioned video backbone $\mathcal{B}$
\State $\{\mathrm{scene}_1, \ldots, \mathrm{scene}_m\} \gets$ decompose $\Sigma$ into \pace{} scene documents conforming to $\mathcal{S}$
\State update $\mathcal{R}_c, \mathcal{R}_p, \mathcal{R}_\ell$ from LLM-assisted character, prop, and location extraction
\For{each shot/panel $u$ in the \pace{} scene documents}
  \State $\mathrm{doc}_u \gets$ resolve \pace{} inheritance for $u$ \Comment{scene defaults $\to$ shot $\to$ panel $\to$ compile hints}
  \State validate PACE bindings in $\mathrm{doc}_u$ \Comment{cast, age states, props, location, compiler-read appearance fields, assets}
  \State $\mathrm{prompt}_u \gets \mathcal{K}_{\mathrm{backend}}(\mathrm{doc}_u)$ \Comment{model dialect only; no change to authored intent}
  \State $\mathrm{keyframe}_u \gets$ render $\mathrm{prompt}_u$ \Comment{appearance anchor from the same bound document}
  \State $\mathrm{scene3D}_u \gets$ assemble Setup, Characters, Props, and Location from $\mathrm{doc}_u$ into a metric scene
  \State $C_u \gets \textsc{PlaceCamera}(\mathrm{doc}_u.\mathrm{camera},\ \mathrm{doc}_u.\mathrm{subjects})$ \Comment{shot size, lens and angle fix position; the aim is solved below}
  \State $\Gamma_u \gets \textsc{CompileTrajectory}(\mathrm{doc}_u.\mathrm{camera.trajectory}, C_u)$
  \State $B_u \gets \textsc{CompileBodySeed}(\mathrm{doc}_u.\mathrm{subjects},\ \mathrm{doc}_u.\mathrm{events})$ \Comment{SMPL-X/Motion-X 322-D frames}
  \State $\mathrm{control}_u \gets$ render the greybox beauty pass plus depth, normal, matte, and object-index passes from $\mathrm{scene3D}_u$ and $B_u$ along $\Gamma_u$ \Comment{the greybox beauty pass is a flat, unstyled render: shape only, no appearance}
  \State $\mathrm{panel}_u \gets \mathcal{K}_{\mathrm{backend}}^{\mathrm{struct}}(\mathrm{control}_u,\ \mathrm{keyframe}_u,\ \mathrm{prompt}_u)$ \Comment{the delivered storyboard still: control\_u's geometry conditions the render, keyframe\_u/reference mattes condition per-subject identity, prompt\_u supplies style}
  \State $\mathrm{clip}_u \gets \mathcal{B}(\mathrm{control}_u,\ \mathrm{doc}_u.\mathrm{resolved\_references},\ \mathrm{prompt}_u)$ \Comment{optional: previs video, a separate artifact from panel\_u}
  \State log artifacts with the \pace{} manifest/provenance fields for $u$
\EndFor
\Statex
\Procedure{PlaceCamera}{$\mathrm{doc}_u.\mathrm{camera}$, $\mathrm{doc}_u.\mathrm{subjects}$}
  \Statex \hskip\algorithmicindent \emph{Resolved \pace{} panel $\mathrm{doc}_u$ with Camera intrinsics/extrinsics, subject world position $\mathbf{p}$ from Characters/Identity \& Staging, and subject screen-position target $(t_x,t_y)$; iteration cap $N$, tolerance $\varepsilon_{\mathrm{tol}}$}
  \State $(\mathbf{c}, \mathbf{r}_0, f) \gets$ camera position, base rotation, and lens resolved from $\mathrm{doc}_u.\mathrm{camera}$
  \State $\mathbf{r} \gets \mathbf{r}_0$
  \For{$i = 1, \ldots, N$}
  \State $(\hat{x}, \hat{y}) \gets \textsc{ProjectToScreen}(\mathbf{c}, \mathbf{r}, f, \mathbf{p})$ \Comment{pinhole projection via the shared rotation matrix}
  \If{$\mathbf{p}$ is behind the camera at $\mathbf{r}$}
  \State \textbf{return} unconverged \Comment{surfaced, not silently accepted}
  \EndIf
  \State $(e_x, e_y) \gets (t_x - \hat{x},\ t_y - \hat{y})$
  \If{$\max(|e_x|, |e_y|) \le \varepsilon_{\mathrm{tol}}$}
  \State \textbf{return} $\mathbf{r}$, converged \Comment{typically $i=1$ for an on-axis subject}
  \EndIf
  \State $(\Delta_{\mathrm{yaw}}, \Delta_{\mathrm{pitch}}) \gets$ angular delta from $(e_x, e_y)$ at the subject's current depth and field of view
  \State $\mathbf{r} \gets \mathbf{r} + (\Delta_{\mathrm{pitch}}, 0, \Delta_{\mathrm{yaw}})$ \Comment{roll held fixed throughout}
  \EndFor
  \State \textbf{return} $\mathbf{r}$, unconverged \Comment{iteration cap reached; reported, not hidden}
\EndProcedure
\Ensure Panels, and optionally clips, whose render inputs remain traceable to their PACE fields
\end{algorithmic}
\end{algorithm}

\subsection{Composition: the read point and the camera aim}
\label{sec:composition}
\label{sec:screen-position-meaning}

\pace{}'s screen-position field needs a fixed referent, because the obvious readings are not equivalent. ``Where the subject sits in the frame'' can mean the centroid of its silhouette or the point the eye lands on. On a dialogue interior those are different places, the torso and the face, and a stage that solves for one satisfies the other only by accident. Measured on the same three rendered panels, the two readings disagree in sign and rank the panels in opposite orders; the silhouette reading gives its best score to the panel whose subject is cut off at the jaw (Section~\ref{sec:composition-precision}). P3 settles the question (Section~\ref{sec:principles}), and \pace{} defines the field at the read point.

\pace{}'s screen-position field is its composition control. It takes a thirteen-option rule-of-thirds zone or a continuous normalized $(x,y)$, plus a depth layer. In the Automatic Drive run, all 45 panels carry a resolvable composition target, and 42 bind that target to cast with screen-position data. That zone is derived, not authored: it follows the subject's order in the cast list, and a fixed zone table maps the three zones used onto $x = 0.38$, $0.50$ and $0.62$ at $y = 0.52$ (Section~\ref{sec:field-provenance}). Honoring the field still takes a real camera aim against project Library geometry rather than the centered default.

Corpus-trained camera models learn this placement, and Storyboard Pro has no field for it at all (Section~\ref{sec:related}). \pace{} neither learns this placement nor leaves it unstated. Given a camera, a lens and a subject's position in an explicit 3D scene, where that subject lands in frame is already determined. The aim that puts it on a specified target is therefore recoverable by a solve over known scene quantities rather than estimated from a corpus (Section~\ref{sec:solver}). What corpus-trained systems obtain through annotation, an explicit scene provides by construction.

The mechanism moves the camera, not the subject, because a subject's world position is shared staging (Section~\ref{sec:blocking}). It inverts the same pinhole model the depth control signal already renders through. \pace{}'s shot-size and angle fields fix the camera position and lens, and the subject's world position comes from the same 3D scene the trajectory renders into. Given these, \textsc{PlaceCamera} (Algorithm~\ref{alg:pace-pipeline}) solves for the pan and tilt, with roll held fixed, that place the subject's projection at the target coordinate; a zone is first resolved to a coordinate through the zone table. Section~\ref{sec:solver} states the solve and its convergence report. The rotation it returns is built by the same rotation construction the camera compiler uses, so a composed shot's camera and its trajectory keyframes are consistent by construction.

\subsection{Blocking and shared staging}
\label{sec:blocking}

Blocking is the second half of composition. Where the camera aims decides what the frame contains; blocking decides how the subjects sit within it. Storyboard Pro leaves it implicit in what an artist draws, with posing deferred to a downstream animation tool (Section~\ref{sec:related}). \pace{} carries it as two per-character fields. \emph{Gaze}, a four-option target type plus a ten-option direction, is compiled into the keyframe prompt as text (grounding tier~1). It stays there because gaze is a direction the model renders an expression toward, not a screen-space coordinate a camera can be aimed to satisfy. \emph{Screen position}, the quantity \textsc{PlaceCamera} solves for, sits at tier~2, because screen placement becomes a function of the same staged 3D scene and compiled camera the trajectory renders from.

A shot's subject position is shared physical staging: the same character stands in the same place across every setup covering it. Composition therefore comes from how the camera frames that fixed position, as a two-shot and its reverse angle compose the same actors differently by moving the camera. Placement and trajectory are then checked against one geometric source of truth instead of being decided independently, as a 2D panel and a separate camera note would be.

\subsection{The aiming solve}
\label{sec:solver}

The claim that composition is computable rather than learned rests on a solve, which we state here. Write the camera's pose as a position $\mathbf{c}$ and a rotation $R$, whose columns are the orthonormal axes $(\mathbf{e}_r, \mathbf{e}_u, \mathbf{e}_f)$ (right, up, forward). For a subject point $\mathbf{p}$, let $\mathbf{v} = \mathbf{p} - \mathbf{c}$ and $d = \mathbf{v}\cdot\mathbf{f}$. The point is in front of the camera when $d > 0$, and its normalized screen coordinates under a pinhole model are
\[
x \;=\; \tfrac12 + \tfrac12\,\frac{\mathbf{v}\cdot\mathbf{e}_r}{d\,\tau_h},
\qquad
y \;=\; \tfrac12 - \tfrac12\,\frac{\mathbf{v}\cdot\mathbf{e}_u}{d\,\tau_v},
\]
with $\tau_h = \tan(\theta_h/2) = w_{\mathrm{sensor}} / (2 f_{\mathrm{mm}})$ the tangent of the half field of view and $\tau_v = \tau_h / \alpha$ for aspect ratio $\alpha$. Given a target $(x^\star, y^\star)$ from the screen-position field, the residual $(\epsilon_x, \epsilon_y) = (x^\star - x,\, y^\star - y)$ maps to an angular correction
\[
\Delta_{\mathrm{yaw}} = \arctan\!\left(2\,\epsilon_x\,\tau_h\right),
\qquad
\Delta_{\mathrm{pitch}} = \arctan\!\left(2\,\epsilon_y\,\tau_v\right),
\]
because a normalized error $\epsilon_x$ corresponds to a lateral offset of $2\epsilon_x\tau_h d$ in the image plane at depth $d$, and the $d$ cancels. Roll is fixed by convention and the focal length comes from the same field group, so only two angles are free and the constraints are exactly determined.

\textbf{The correction is closed form; the solve is not.} Each correction is a closed-form expression, but applying it changes $R$, which changes the projection, so the solve is a fixed-point iteration rather than a single inversion. It runs to a tolerance of $10^{-5}$ in normalized screen units within a cap of 16 iterations, and reports non-convergence rather than returning a pose that misses the target. Two things prevent a one-shot solution. The projection is nonlinear in the rotation, so the correction is exact only in the small-residual limit. And when the target is the midpoint of a subject's head and foot projections, the driven quantity itself depends on the rotation. That mode needs the larger cap; eight steps suffice for a single point.

The claim is therefore not that camera aiming admits a one-line analytic solution in general, but that it needs no learned prior. The residual is computable from the scene at every step, so the solve is a numerical procedure over known quantities rather than an estimate from a corpus. Its failure conditions are geometric and reportable: a subject behind the camera, or a target that the lens cannot reach at that pose.

\subsection{Compiling camera movement: language to a metric track}
\label{sec:trajectory}

\textsc{CompileTrajectory} (Algorithm~\ref{alg:pace-pipeline}) moves the camera that the aiming solve places. It shares the static solve's first steps: subjects resolved to world positions, a lens and starting distance from the declared shot size, and a starting pose from the declared angle and relative position. Where the movement is authored in free text, an off-the-shelf LLM classifies it against a small library of named camera moves, each carrying its own evidence grade and citation. A move for which the library has no textual support is left unclassified. The matched move's primitives, each with a duration and an easing curve, compile deterministically into a metric track of one position, rotation and lens value per frame, aimed throughout at the shot's read point. The convergence and behind-camera checks of Algorithm~\ref{alg:composition-solve} apply at every frame.

No panel of the evaluated corpus declares a moving camera (Section~\ref{sec:camera-adherence}), so the corpus results say nothing about this compiler.
Two demonstrations stand in for the evidence the corpus cannot give.

\textbf{A worked move.} The dramatic function \emph{tension} in the move library
is sourced to the convention that a slow push tightens the frame around a
subject as pressure mounts. Its one skill is a push-in eased in rather than
linear, so the move starts slow and gathers speed. Compiled to an 8-frame
segment at a 35\,mm lens, starting 6\,m back and closing to 3\,m, it produces
the track in Table~\ref{tab:push-in-track}, which is compiler output rather
than a hypothetical.

\begin{table}[tbp]
\centering
\footnotesize
\caption{An eased push-in compiled to a metric track: the per-frame change in
distance grows as the move gathers speed, and the pitch steepens to hold the
same world read point, 90 degrees being level. Eight frames at a 35\,mm lens,
from 6\,m to 3\,m.}
\label{tab:push-in-track}
\begin{tabular}{@{}rrrr@{}}
\toprule
frame & distance (m) & pitch (deg) & per-frame $\Delta$distance \\
\midrule
0 & 6.00 & 76.87 & -- \\
1 & 5.94 & 76.74 & 0.06 \\
2 & 5.76 & 76.33 & 0.18 \\
3 & 5.45 & 75.59 & 0.31 \\
4 & 5.02 & 74.42 & 0.43 \\
5 & 4.47 & 72.61 & 0.55 \\
6 & 3.80 & 69.76 & 0.67 \\
7 & 3.00 & 64.98 & 0.80 \\
\bottomrule
\end{tabular}
\end{table}

The growing per-frame distance column is the ease-in curve made concrete. The
pitch column moves too, although nothing asked it to: holding the same world
read point at a shrinking distance requires a steeper downward tilt, from 13
degrees below level to 25. This is the coupling Section~\ref{sec:solver} states
for a static pose, applied once per frame along the move. Frames 0 and 7 are
the start and end framings a moving shot would hand to the densification step,
which today adds the second panel without carrying either endpoint's camera
pose onto it (Section~\ref{sec:breakdown-eval}).

\begin{figure}[!tbp]
  \centering
  \includegraphics[width=0.98\textwidth]{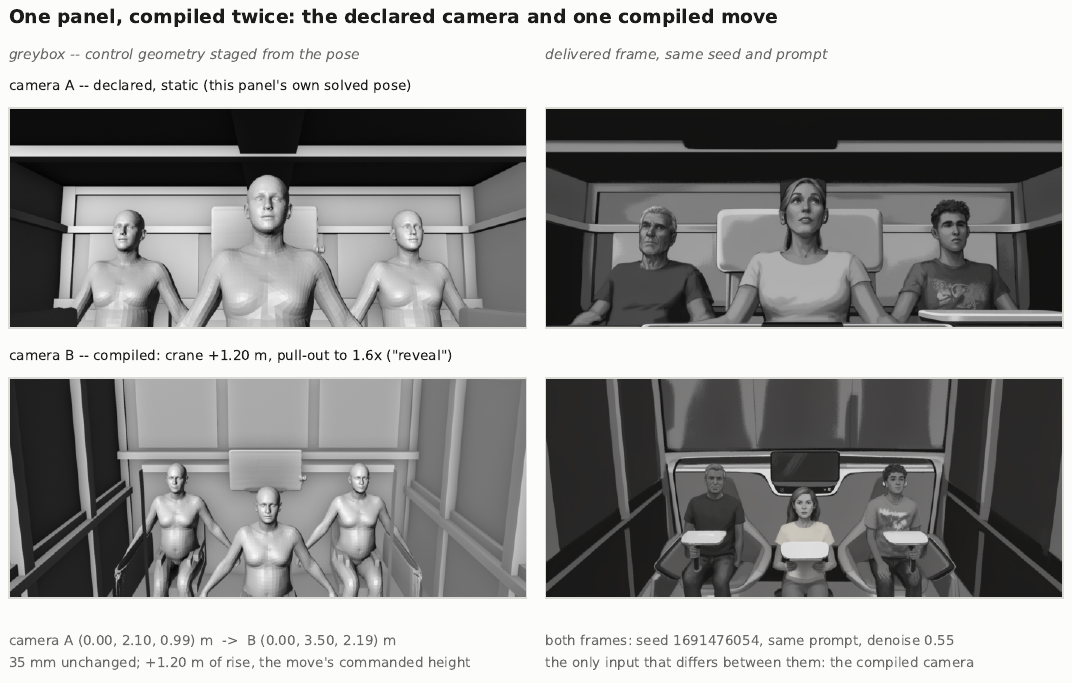}
  \caption{Compiling one move onto a real panel changes the staged camera and
  nothing else. \textbf{Top row:} camera A, the panel's own solved static pose,
  unchanged. \textbf{Bottom row:} camera B, the same pose with one compiled move
  applied. \textbf{Left column:} the control geometry each camera stages.
  \textbf{Right column:} the delivered frame, with the same seed and compiled
  prompt as the panel's own corpus delivery in both rows, so the pair isolates
  the compiled camera as the only variable. This is a demonstration outside the
  corpus's own adherence claims (Section~\ref{sec:camera-adherence}): the shot
  does not declare a moving camera, and the figure shows what compiling one for
  it produces.}
  \label{fig:camera-movement}
\end{figure}

\textbf{One compiled move on a real panel.} Running the stages end to end once,
outside the corpus's own adherence claims, fills the gap for one real panel,
\texttt{scene\_03/shot\_01/panel\_0001}. Its own solved static pose, from the
same fit-to-cast solve every panel in the corpus gets, is camera A. Camera B
compiles the library's \emph{reveal} move against it: a rise of 1.20\,m
followed by a pull-out to 1.6\texttimes{} the starting distance, sourced to
standing craft usage rather than a citation. Both cameras stage the same cast,
and both control images render through the same workflow, base model and seed
as the panel's own corpus delivery, so the only input that differs between the
two delivered frames of Figure~\ref{fig:camera-movement} is the compiled
camera. Its world position moves from $(0.00, 2.10, 0.99)$\,m to $(0.00, 3.50,
2.19)$\,m, and the $+1.20$\,m of rise matches the move's own commanded height
exactly.

The pair also shows a limit. The compiled prompt reads ``Captured as an extreme
close-up'' in both frames, because shot size compiles into prompt text from a
declared field the movement does not touch. The geometry moves and the words
describing it do not, a second instance of the translation gap
Section~\ref{sec:resolved-not-used} reports.

\label{sec:events-smplx}
\textbf{Grounding the action, emotion and dialogue fields in a parametric body.} These three schemas, which \scine{} files under Events and \pace{} gathers under Characters (Section~\ref{sec:schema}), are the largest block of fields left at tier~1, because a greybox of primitives can stage where a body is but not that it runs, flinches or speaks. SMPL-X~\cite{smplx2019} matches the three schemas one for one (body pose, facial expression, jaw), and the handoff already writes per-subject 30-frame seeds in the 322-dimensional SMPL-X/Motion-X convention, from coarse presets rather than fitted to the authored action. Fitted performance is 0 of 76 subject vectors (Section~\ref{sec:adherence-experiment}).

\section{Evaluation}
\label{sec:evaluation}

The measurements are layered, not reduced to one score, because a representation, a renderer and a fitting path can each fail independently, and an aggregate would hide which one did. They follow the three implemented controls in the order a panel acquires
them (breakdown, composition, and camera), then analyze continuity across
existing cuts without treating the unimplemented transition field as a
delivered control, and then test the configured workflow against prompt-only
arms on an external corpus. A final group bounds what the rest of the paper may claim. Table~\ref{tab:results-summary} lists the principal measurement of each experiment with the level it is taken on. \emph{Automatic Drive} holds 11 scenes, 27 shots and 45 panels, 42 of them with cast; the composition step of Section~\ref{sec:shot-size-adherence} split six group panels into fifteen singles, so 45 = 36 - 6 + 15, and experiments run before that step report 36 panels and say so. The same cast counts 58 declarations at shot level and 76 references at panel level, and panels in film order give 34 within-scene cuts. One experiment leaves this corpus: a baseline on PROSE, an external set of director storyboards (Section~\ref{sec:prose-baseline}). Every render reported here was produced on one shared machine using four RTX 5090s, by Flux~2 dev in fp8 at $1280\times544$, at seeds 101, 202 and 303 unless a figure names its own, with the greybox staged in Blender and the specification compiled by \texttt{pace-core}; no hosted service is in the loop.

\begin{table}[!tbp]
\centering
\footnotesize
\caption{The representation checks pass on every panel and the camera checks on
all but the over-the-shoulder's angle, while
fitted motion, multi-subject placement, pairwise overlap and the gate's
occlusion clause report measured failures; each row is one experiment's
principal measurement. The line under each count names the level measured
(the record, the staged geometry by projection, the rendered staging by
object-index mattes, or the generated image), and \emph{external} marks an
instrument not derived from the specification. There is no aggregate to fold a
failure into.}
\label{tab:results-summary}
\newcommand*{\instr}[1]{\newline{\itshape\color{black!58}#1}}%
\newcommand*{\group}[2]{\multicolumn{4}{@{}l}{\footnotesize\textbf{#1}: #2}}%
\scriptsize
\setlength{\extrarowheight}{1.5pt}%
\begin{tabularx}{\textwidth}{@{}R{4.5cm}R{2.3cm}Xr@{}}
\toprule
Experiment\instr{instrument} & Over\instr{level} & Result & Sec. \\
\midrule
\group{Control one}{script breakdown} \\
Artifact-chain availability\instr{compiled artifacts present and bound} & 45 panels\instr{record} & Four representation criteria pass on every panel; fitted motion fails on every scorable subject; visual review is unmeasured & \ref{sec:adherence-experiment} \\
Adjacent-panel distinguishability\instr{field-by-field diff of sibling panels} & 9 densified shots\instr{record} & Neither panel of a densified shot differs from its sibling on a camera field & \ref{sec:breakdown-eval} \\
\group{Control two}{composition at exact object positions} \\
Placement of the focus subject\instr{head-mask centroid vs.\ declared $x$} & 42 panels\instr{rendered staging} & 36 of 42 within 5\% of frame width, mean 2.1\%, worst 10.6\% on the over-the-shoulder; every single-subject panel within 1.2\% & \ref{sec:composition-precision} \\
Positions under the fixed-staging solve\instr{projected position vs.\ declared $x$} & 76 placements\instr{staged geometry} & Single subject 0.3\% mean, 1.1\% max; all 76: mean 4.0\%, 61 within 5\%; multi-subject targets retain shared residual & \ref{sec:placement-budget} \\
Placement on the generated image\instr{pixel proxy, no segmentation} & 33 panels\instr{generated image} & Across the style repair, mean error 9.4\% $\to$ 4.7\% of frame width, worst 28.1\% $\to$ 15.3\% & \ref{sec:visual-adherence} \\
\group{Control three}{camera} \\
Angle, position and lens\instr{view elevation, azimuth, lens} & 42 panels\instr{compiled camera} & Lens and position 42/42; angle 40/42, the two over-the-shoulder panels at $-1.4^\circ$; no moving camera & \ref{sec:camera-adherence} \\
Delivered framing vs.\ shot size\instr{body height as share of frame} & 42 panels\instr{rendered staging} & Where height decides the distance (35 of 42), every size within ten points of its target, e.g.\ \texttt{wide} 73.3\% against 76.9\%, \texttt{extreme\_close\_up} 87.5\% against 94.3\% & \ref{sec:shot-size-adherence} \\
Frame containment of the cast\instr{matte contact with each frame edge} & 76 placements\instr{rendered staging} & 44 of 76 clipped at the bottom as the shot size intends; 2 touch a side, both an over-the-shoulder's near shoulder & \ref{sec:frame-containment} \\
\group{Cut analysis}{continuity across existing cuts; transition field unimplemented} \\
Staging drift across a cut\instr{staging delta vs.\ declaration delta} & 34 in-scene cuts\instr{staging} & 12 unchanged cuts held within 1\%; staging moves on 6 of 8 changes; 14 have no subject on both sides; the lens moves at 20 and turns at 2 & \ref{sec:transition-adherence} \\
Screen direction\instr{matte order and side of center} & 32 pairs\instr{rendered staging} & 32/32 pairs hold left-to-right order; 44 of 44 subject crossings stay on their own side & \ref{sec:screen-direction} \\
Style continuity\instr{flat-region share} & 33 panels\instr{generated image} & Absolute median cut step 2.5 $\to$ 5.2 points; normalized step 14.4\% $\to$ 9.0\%; neither crosses the uncalibrated threshold & \ref{sec:look-continuity} \\
\bottomrule
\end{tabularx}
\end{table}

\begin{table}[!tbp]
\ContinuedFloat
\centering
\caption{Principal measurements continued: the external baseline, and
diagnostics that bound the scope of the three implemented controls and the cut
analysis.}
\newcommand*{\instr}[1]{\newline{\itshape\color{black!58}#1}}%
\newcommand*{\group}[2]{\multicolumn{4}{@{}l}{\footnotesize\textbf{#1}: #2}}%
\scriptsize
\setlength{\extrarowheight}{1.5pt}%
\begin{tabularx}{\textwidth}{@{}R{4.5cm}R{2.3cm}Xr@{}}
\toprule
Experiment\instr{instrument} & Over\instr{level} & Result & Sec. \\
\midrule
\group{External baseline}{prompt-only arms on PROSE director storyboards} \\
Staged checks, external corpus\instr{staged geometry vs.\ declared fields} & 204 PROSE panels\instr{staged geometry} & Declared size realized on 204/204; single-subject placement mean 0.12\%, max 0.17\%; two-subject mean 8.0\%; left-to-right order 49/49 & \ref{sec:prose-baseline} \\
Prompt-only baselines\instr{head height vs.\ staged; VLM judge} & 204 shots, 2448 renders\instr{generated image; external} & Heads come back at 1.906$\times$ their staged height from the director's own words and 1.733$\times$ from the compiled prompt, against 0.955$\times$ with the greybox; size order holds in 82.0\%, 90.0\% and 97.1\% of pairs. Described content runs the other way, seen in 87.1\%, 67.0\% and 53.9\% of renders and actions in 96.2\%, 76.7\% and 57.5\%, by an unvalidated judge. Declaring a pose recovers part of the last figure: on the 30 shots it moved, actions rise to 74.4\% from 58.9\% at an unchanged 0.960$\times$ staged height & \ref{sec:prose-baseline} \\
\group{Beyond the primary controls}{scope bounds and diagnostics} \\
Entity consistency vs.\ gap\instr{whole-frame similarity by shot gap} & 36 panels, pre-split\instr{generated image; external} & No measurable identity signal once location change is held constant; the uncontrolled decay is an artifact & \ref{sec:entity-consistency} \\
Provenance of declared fields\instr{value entropy per leaf path} & 130 fields, 3267 instances\instr{record} & 38 take one value; beyond 14 constant by nature in one film and 19 in groups the corpus does not exercise, the finding is five fields, screen position the worked case & \ref{sec:field-provenance} \\
Grounding against the screenplay\instr{set difference vs.\ extracted events} & 79 events\instr{record; external} & All 79 land in exactly one beat; 22 state transitions survive segmentation & \ref{sec:verified-breakdown} \\
Estimator vs.\ planted errors\instr{recall and precision against planted errors} & 18 mutations\instr{external (LLM)} & Strict recall 0.71 over 14 measurable mutations and precision 0.37; a deleted event is re-attached rather than reported missing & \ref{sec:verified-breakdown} \\
Gate between anchor and diffusion model\instr{clause margins} & 42 panels\instr{staged geometry} & 27 pass; 11 fail on occluded rear cabin passengers, 5 on a frame edge at a joint; reports at build time, refuses no render & \ref{sec:greybox-gate} \\
Two delivered-pixel failure modes\instr{compiled prompt vs.\ delivered image} & 33 panels, pre-split\instr{generated image} & A beat-state contradiction on 2 of 27 shots, addressed by a per-shot state channel; one style field delivering two styles, addressed through the control image & \ref{sec:visual-adherence} \\
Identity plates, ablated\instr{head height vs.\ staged} & 42 panels, 84 renders\instr{generated image} & Attaching the plates leaves the staging where it was: delivered head height 0.976$\times$ staged without them against 0.968$\times$ with, 22 of 42 panels closer with them ($p = 0.88$); declared cast recovered on 34 and 36 panels ($p = 0.63$); what they buy is not measured & \ref{sec:refplate-ablation} \\
\bottomrule
\end{tabularx}
\end{table}

\subsection{Specification-derived conformance audit}
\label{sec:self-oracle}

The specification provides a test oracle for compiled and staged artifacts, not
for the generated panel as a whole. Fields are typed and drawn from closed
vocabularies or bounded numbers; tier-2 fields are discharged into a staged
metric scene; and the staging renders one object-index matte per declared
subject. These properties make camera and geometric residuals measurable
without a learned evaluator because subject identity is known on the staged
passes by construction.

Once the diffusion model produces a delivered panel, those mattes are no longer
pixel ownership labels for the generated image. Rows measured on delivered
pixels therefore name a proxy, detector, embedding model, language-model judge,
or human review explicitly. The audit also checks conformance to the
declaration, never whether the declaration was a good directing decision.
Section~\ref{sec:composition-precision} gives a case where a plausible metric made the frame worse.

\subsection{Control one: script breakdown}

The breakdown decides where one panel ends and the next begins, so it is measured first: how much of a real screenplay survives the path, and whether each panel earns its place under P1.

\subsubsection{Artifact-chain availability on Automatic Drive}
\label{sec:adherence-experiment}

We ran the handoff path on the public Automatic Drive short-film script and scored it with layered criteria, because final visual quality and \pace{} adherence are different measurements. Four criteria test the representation: structural coverage (a compiled prompt, rendered PNG, applicable SMPL-X/Motion-X seed and camera work item exist), \pace{}-field adherence (event delta, typed specification bindings and resolved read point are present), asset-anchor adherence (visible entities resolve to the project Library) and camera-intent adherence (a ready camera work item carries the target read point). Two test the output: fitted-motion adherence (character panels carry fitted, not coarse, SMPL-X/Motion-X performance) and final storyboard visual adherence under human review. The four representation criteria pass on every panel at both sizes of the corpus, the 27-panel cut and the current 45 panels: 45/45 prompts compiled, bindings and read points present, and camera items ready with none blocked. One output criterion fails on every scorable subject, fitted performance being 0 of 76 subject vectors (0 of 24 panels on the smaller cut), because the artifact is a coarse parameter set; the other is unmeasured, since a human review of one set of images is not evidence about another. \pace{} therefore localizes the failure, to the renderer, the reference conditioning and the motion-fitting path; it does not eliminate it.

Coverage bounds every later measurement, since a field left empty here cannot be recovered downstream. Shot size and camera fill 45/45 panels but are a director's authored shot design, since the screenplay names no camera term; cast, which \emph{is} extracted, reaches 42/45, the three gaps falling in the one scene the breakdown left empty. Every subject in the 42 cast panels has a screen position, assigned by list order through a zone table, not authored per subject (Section~\ref{sec:field-provenance}). Referential integrity is complete: all 76 character references resolve and name a likeness asset, and no age-state, prop or location reference dangles. Segmentation obeys P1 at the shot level, 0 of 27 shots failing any of its four rules (Section~\ref{sec:beat-gates}), but not yet at the panel level.

\subsubsection{Adjacent-panel distinguishability after densification}
\label{sec:breakdown-eval}

Neither panel of a densified shot differs from its sibling on a camera field. Densification adds a second panel to a shot that needs a second framing target; nine of the 27 shots received one. On these nine the schema's permission is not yet used correctly: neither panel carries a \texttt{camera\_override} or a \texttt{setup\_override}, so what distinguishes the rendered pair is decided downstream of the representation, and 9 of 44 adjacent panel pairs compile to the same camera, staging and lighting, eight of them densified pairs. This is reported and not corrected, because what a densified panel's camera should say is a directorial decision.

\subsection{Control two: composition at exact object positions}

Composition is measured at two levels, and each result names its level: where the focus subject lands, on the rendered staging with an object-index mask per subject, and how many declared positions one camera pose can hold, on the staged geometry by projection. Neither is the generated image, whose placement is a pixel proxy reported in Section~\ref{sec:visual-adherence}.

\subsubsection{Composition precision on the rendered staging}
\label{sec:composition-precision}

On the rendered staging, the focus subject's head-mask centroid lands within 5\% of frame width of its declared horizontal target on 36 of 42 panels, with a mean of 2.1\% and a worst case of 10.6\%, and every single-subject panel lands within 1.2\% (19 of 19, mean 0.35\%). Four of the six misses are multi-subject panels whose cast cannot all be held in one frame: two at 8.8\% and 8.7\% (Section~\ref{sec:placement-budget}), and scene~11's two-shot at 6.0\% on each of its two panels. The other two are the over-the-shoulder, at 10.5\% and 10.6\%, and they miss for a structural reason rather than a solver one: screen position does double duty, setting a subject's seat in the scene and naming where the frame should put them, and this is the one position that honours neither the usual way. Its lens is solved from the pair, so the far subject lands near centre, at 0.485 or 0.514, whatever $x$ is declared, and the pair is not jointly satisfiable. The check is needed because the solver's pinhole model is an approximation, one aim point per subject and not a full-body volume. It uses the segmentation pass every production render already carries, and P3 selects the head over the whole-body silhouette (Section~\ref{sec:principles}). The choice matters: on a three-panel pilot the two readings disagreed in sign and ranked the panels in opposite orders, and aiming at a whole-body midpoint drove the silhouette error below 3\% while tilting the head out of frame.

A check on rendered pixels catches what projection cannot: a wrong convention, such as the sign of world $+X$ under a backward-facing cabin camera, projects consistently and still puts a subject on the wrong side. Grounding a value in geometry checks that it is used consistently, not that its convention is the intended one. The precision is also only as informative as the targets it is measured against. The corpus declares three screen-$x$ patterns across its 24 multi-subject shots and one $y$ for all 58 subjects, so placement error against these constants describes the solver, not the blocking.

\subsubsection{Positions satisfiable by one camera pose}
\label{sec:placement-budget}

\begin{figure}[!tbp]
  \centering
  \begin{tikzpicture}
    \begin{axis}[
      width=0.66\textwidth, height=5.6cm,
      xlabel={declared screen $x$}, ylabel={measured screen $x$},
      xtick={0.38,0.50,0.62}, xmin=0.30, xmax=0.70,
      ytick={0.1,0.38,0.5,0.62,0.9}, ymin=0.05, ymax=0.95,
      tick label style={font=\scriptsize}, label style={font=\scriptsize},
      legend style={font=\scriptsize, draw=none, fill=none, at={(0.02,0.98)}, anchor=north west, cells={anchor=west}},
      axis line style={pacegrid}, tick style={pacegrid},
      grid=none, clip=false,
    ]
      % the target: a placement that landed where it was declared sits on this line
      \addplot[pacemuted, dashed, thin, domain=0.30:0.70, forget plot] {x};
      \node[font=\tiny, text=pacemuted, anchor=south west, rotate=27] at (axis cs:0.635,0.635) {declared = measured};
      % single-subject panels: the solve has one target and hits it
      \addplot[only marks, mark=square*, mark size=1.9pt, paceink,
               x filter/.expression={x-0.014}]
        coordinates {(0.50,0.503) (0.38,0.391) (0.62,0.621) (0.50,0.503) (0.38,0.391) (0.62,0.621) (0.50,0.498) (0.38,0.383) (0.62,0.618) (0.50,0.500) (0.38,0.387) (0.50,0.497) (0.38,0.385) (0.50,0.497) (0.38,0.385) (0.50,0.500) (0.50,0.500) (0.50,0.497) (0.50,0.497)};
      \addlegendentry{single subject (19)}
      % the focus subject of a multi-subject panel: the aim shares the residual with it
      \addplot[only marks, mark=*, mark size=1.8pt, paceblue, fill=paceblue!85]
        coordinates {(0.50,0.500) (0.50,0.500) (0.50,0.500) (0.50,0.500) (0.50,0.535) (0.50,0.535) (0.50,0.588) (0.50,0.536) (0.50,0.493) (0.50,0.493) (0.50,0.503) (0.50,0.503) (0.50,0.413) (0.50,0.493) (0.50,0.493) (0.50,0.455) (0.50,0.455) (0.50,0.455) (0.50,0.455) (0.62,0.680) (0.62,0.680) (0.38,0.485) (0.62,0.514)};
      \addlegendentry{focus subject, multi-subject panel (23)}
      % the other subjects: placed by the staging, not by the aim
      \addplot[only marks, mark=triangle*, mark size=2.1pt, paceorange, fill=paceorange!85,
               x filter/.expression={x+0.014}]
        coordinates {(0.38,0.338) (0.62,0.662) (0.38,0.338) (0.62,0.662) (0.38,0.338) (0.62,0.662) (0.38,0.338) (0.62,0.662) (0.38,0.337) (0.62,0.584) (0.38,0.336) (0.62,0.585) (0.38,0.280) (0.62,0.645) (0.38,0.331) (0.62,0.587) (0.38,0.387) (0.38,0.387) (0.38,0.382) (0.62,0.701) (0.38,0.382) (0.62,0.701) (0.38,0.221) (0.62,0.859) (0.38,0.387) (0.38,0.387) (0.38,0.424) (0.38,0.424) (0.38,0.424) (0.38,0.424) (0.38,0.319) (0.38,0.319) (0.62,0.906) (0.38,0.093)};
      \addlegendentry{non-focus subject (34)}
    \end{axis}
  \end{tikzpicture}
  \caption{A single subject lands on its declared position wherever it is
  declared, while a panel declaring several cannot hold them all and the
  arbitration of Eq.~\ref{eq:arbitration}, at $w = 0.5$, spreads what is left
  over rather than parking it on whoever is not the read point. Each of the 76 marks over 42
  panels plots a subject's declared screen $x$ against where the staged
  geometry projects it (staged geometry, not the generated image); the dashed
  line is the target. The focus subject lands between 0.413 and 0.588 for a
  declared 0.500, and the others drift outward from center on both sides;
  the marks furthest from the line, at 0.093 and 0.906, are the near
  shoulders of a shot / reverse-shot, whose lens is solved from the pair,
  consistent with the limited degrees of freedom described in
  Section~\ref{sec:formal}.
  Marks stack where the panels of a scene share a camera; the small
  horizontal offset between series only lets the three be told apart.}
  \label{fig:placement-by-declaration}
\end{figure}
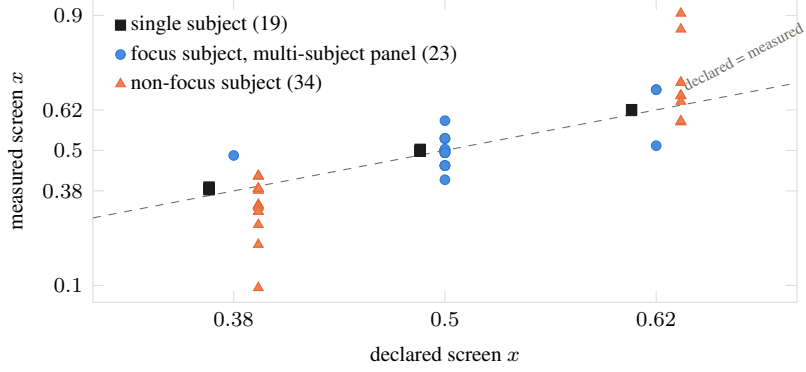

Read off the staged geometry by projection, the 76 subject placements in the 42 staged panels deviate from their declared screen positions by a mean of 4.0\% of frame width (median 3.5\%, maximum 28.7\%); 61 land within 5\% and 70 within 10\%. The error separates on whether a panel declares one subject or several. A single subject is satisfied to 0.3\% on average and 1.1\% at worst, whether it declares the center or a side (Figure~\ref{fig:placement-by-declaration}); two-subject placements average 6.0\% and three-subject placements 4.7\%. The two-subject figure carries scene~11's shot / reverse-shot, whose lens is solved from the pair: it keeps both subjects on their declared sides of 0.38 and 0.62 but frames the face at 0.485 or 0.514 and the near shoulder at the edge, 0.092 or 0.906; without its four placements two subjects average 3.3\% and the maximum is 23.9\%. Declarations of center are realized at 0.413--0.588 against 0.500.

This is a limitation of the current fixed-staging solve, not a general
infeasibility result. It varies two angles while screen position is declared per
subject, so three targets supply six scalar demands to two variables
(Section~\ref{sec:formal}). The implementation arbitrates toward similar errors
because aiming only at the focus puts seven declared subjects of this corpus out
of frame. Fifteen of the 19 single-subject placements are singles split from group
shots, and nine of the 19 keep an off-center target; the 1.1\% maximum holds
across both. Reducing multi-subject residual could instead
free camera position or lens within the shot class, jointly optimize blocking,
or mark secondary positions advisory. The present system keeps staging fixed
and reports the residual for every subject.

On the rendered staging's mattes, ordinal relations between subjects survive and metric ones do not: order holds on 23 of 23 multi-subject panels and depth on 28 of 30 pairs, while spacing widens by 5.7\% of frame width and 12 of 45 pairs overlap by more than 2\%.

\subsection{Control three: camera}

The camera is measured twice: on the compiled pose, against the angle, position and lens the panel declares, and on the rendered staging, against the framing the declared shot size asks for.

\subsubsection{Compiled camera against declared angle, position and lens}
\label{sec:camera-adherence}

The declared lens and position are realized on every panel that compiles a camera, 42 of 42 on both, and the declared angle on 40 of 42. Three declared fields drive the camera, \texttt{angle}, \texttt{position} and \texttt{lens\_mm}, and the build returns the pose it compiled, so the check is on the compiled camera, not on the viewpoint of the generated image; elevation and azimuth are read off the view direction, since one pose has several Euler representations. Across the 42 panels that compile a camera, the focal length equals the declared one on 42 of 42. Declared angle reproduces exactly wherever the position leaves it free: \texttt{high\_angle} at $-14.0^\circ$ on 6 of 6, \texttt{low\_angle} at $+8.0^\circ$ on 6 of 6 and \texttt{eye\_level} at $-5.0^\circ$ on 28 of its 30, correctly ordered. Declared position reproduces on all 42: \texttt{front} at $0^\circ$ off the cabin axis on 36 of 36, \texttt{three\_quarter} at $25^\circ$ on 4 of 4, and the 2 over-the-shoulder panels placed from their pair rather than from an azimuth, which is what that value means. Those 2 are also the angle's two exceptions, and they are one place where camera fields do compete: the lens is pinned to the near subject's shoulder, the far subject's head sits above that shoulder, and holding the declared $-5.0^\circ$ on the far face would put the lens over the near crown, where the frame keeps the angle and stops being an over-the-shoulder. The build reports the shortfall rather than the angle it did not reach, and the panels come out at $-1.4^\circ$. Everywhere else a shot has one camera and its fields do not compete, which is why the solve that cannot satisfy three declared screen positions satisfies these. The corpus exercises three angle values and three of five positions, and \emph{no panel declares a moving camera}, so this data does not exercise the trajectory compiler of Section~\ref{sec:trajectory}.

\begin{figure}[!tbp]
  \centering
  \includegraphics[width=0.98\textwidth]{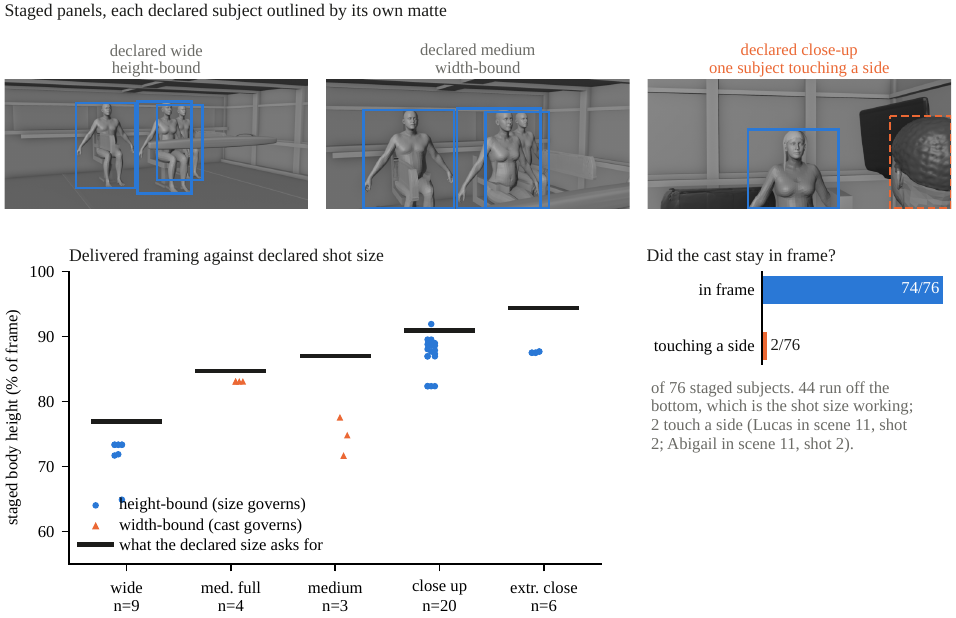}
  \caption{The declared angle, position and lens are realized exactly, while
  the declared shot size is delivered only where the vertical axis decides the
  camera distance. \textbf{Top:} three staged panels, each declared subject
  outlined by its own object-index matte, solid blue when clear of both sides of
  the frame and dashed orange when touching a side. \textbf{Bottom left:} body
  height on the rendered staging against the declared shot size, one marker per
  panel, with the black bar marking what the size asks for; where the vertical
  axis decides the camera distance (circles) the size is delivered, and where
  the group's width decides (triangles) the miss grows with the tightness of
  the declaration (panels of one scene share a camera and coincide, hence the
  counts on the axis). \textbf{Bottom right:}
  what that trade bought, the cast kept in frame. Bodies running off the
  \emph{bottom} of frame are the shot size working and are not counted as loss
  (Section~\ref{sec:frame-containment}).}
  \label{fig:camera-adherence}
\end{figure}

\subsubsection{Delivered framing against declared shot size}
\label{sec:shot-size-adherence}

Where the vertical axis decides the camera distance, which is 35 of the 42 panels, every declared size is delivered within ten points of its target. Shot size is a result, the share of the frame a body occupies, so it is measured on the rendered staging, off each subject's matte. The fit has to clear the cast in both axes and takes whichever distance is larger, so the corpus splits in two (Figure~\ref{fig:camera-adherence}). On the 35 panels where the vertical axis decides, the declared size governs: \texttt{wide} stages at a median 73.3\% of frame height against a target of 76.9\%, \texttt{close\_up} at 88.0\% against 90.9\% and \texttt{extreme\_close\_up} at 87.5\% against 94.3\%, every panel within ten points of its target. On the seven where the group's width decides, the band stops binding on the looser sizes: \texttt{medium\_full} lands at 83.1\% against 84.7\% and \texttt{medium} at 74.8\% against 87.0\%. Two of those three \texttt{medium} panels are the over-the-shoulder, and they are the widest miss in the corpus for a reason the position states outright: its distance is solved from the pair, so the framed subject lands where clearing the near shoulder puts it rather than where the size band asks.

The tight sizes hold because of a composition step before the build. A panel declaring a close-up or tighter on two or more subjects, whose frame is decided by width, is first restaged and, failing that, split into one single per subject, the way a tight exchange is normally covered. On this corpus no restage satisfied the band, so six panels were split into fifteen singles, and the 76 placements are unchanged.

\subsubsection{Frame containment of the declared cast}
\label{sec:frame-containment}

No declared subject is lost from frame. A subject counts as leaving the frame at an edge its body matte touches. Across the 76 staged subjects in 42 panels, every one has a matte and none touches the top edge. Two touch a side, and both are the near shoulder of scene~11's shot / reverse-shot, which an over-the-shoulder places at the frame edge by design. Forty-four touch the bottom edge, and they follow the declared size: a tighter size keeps less of the body, so none of the 21 subjects in \texttt{wide} panels touches the bottom, against 4 of 12 in \texttt{medium\_full}, 7 of 7 in \texttt{medium}, 27 of 30 in \texttt{close\_up} and 6 of 6 in \texttt{extreme\_close\_up}. The distance that misses the size band on the seven width-decided panels of Section~\ref{sec:shot-size-adherence} is the distance that keeps their cast inside the frame.

\subsection{Cut analysis with the transition field unimplemented}

The transitions field is proposed but not implemented, so this control is measured as continuity across the cuts the corpus already contains: what holds when the declaration does not change, what moves when it does, and whether the reader's cue for continuity survives.

\subsubsection{Staging drift across cuts}
\label{sec:transition-adherence}

Staging holds wherever the declaration is unchanged and follows wherever it changes. Each panel is compared with its successor in film order, giving 34 cuts inside the 11 scenes; staged positions, not the generated image, are compared, and two are the same when they agree within 1\% of frame width. The blocking declaration is what places a body: the camera, each subject's screen position and each subject's pose; a change of gaze alone moves no one. Twelve cuts leave that declaration unchanged, and the staging holds on all 12 within 1\%. Eight change it, and the staging moves on six; on the other two the change is a pose that leaves the body where it stood. The other 14 have no subject on both sides: ten are shot / reverse-shot cuts between the singles of a split panel, two join the empty panels of scene~4, and two change the whole cast. Read narrowly, the result says the pipeline adds no drift of its own, and the corpus tests the contract lightly: 4 of 34 cuts change a declared camera value. The solved lens still moves at 20 of them, by 2\,cm to 4.7\,m, because it is fitted again to each panel's cast and shot size, and it turns to a new direction at only two: into scene~11's exchange, and from its shot to the reverse shot, where the panel's focus rather than a camera value turns it around. The stability of the unchanged cuts follows from recomputing the staging deterministically, not from the schema. Where scene~2 cuts in, unchanged screen positions land differently on either side, the cast spreading from 0.251 to 0.370 of frame width, because screen position is realized through the camera.

\subsubsection{Screen direction across the same cuts}
\label{sec:screen-direction}

\begin{figure}[!tbp]
  \centering
  \includegraphics[width=0.98\textwidth]{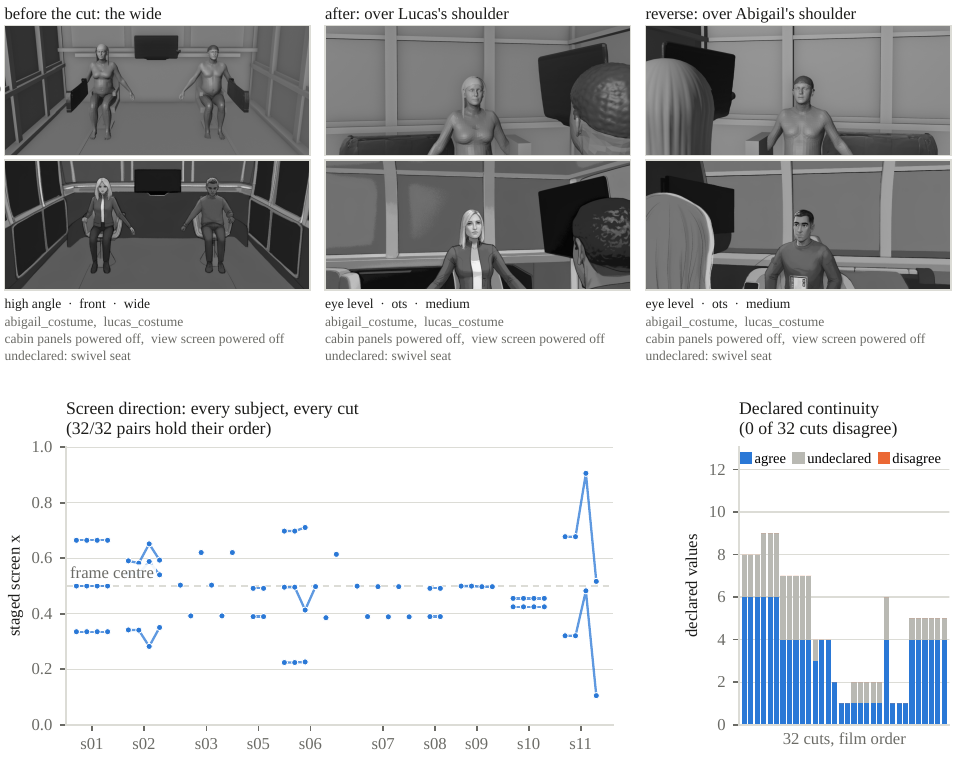}
  \caption{A two-person exchange covered as shot / reverse-shot is built in one cabin and holds its screen direction across the cuts that turn the lens.
  \textbf{Top:} scene~11 cuts from a wide, front-on two-shot to a medium,
  eye-level over-the-shoulder onto Abigail and its reverse onto Lucas, both
  lenses kept on the side of the line the wide sets. All four panels of the
  scene are built inside one cabin, the union of what each camera needs, so
  each reverse angle looks at a side wall the wide shows, with the same bays.
  Each panel is shown staged (upper row) and delivered from one fixed seed
  (lower row), above what it declares: the camera, the wardrobe entries and
  the state of each prop. The greyboxes carry the wardrobe as garment tones,
  and each character's hair as a shell toned from its declared colour;
  shoulder-length hair hangs below the head, so it has the same length seen
  from the front and from behind. At this seed Abigail keeps her jacket and her
  light shoulder-length hair in all three panels, and Lucas his short hair and
  sweatshirt; his hair, declared black flecked with grey, reads grey in the
  wide and dark in the two closer shots. The wide came back in colour against the declared
  greyscale style and is shown in greyscale. The lower row is one render per
  panel, not a selection.
  \textbf{Bottom left:} each subject's staged screen $x$ through the
  film. Every pair of subjects keeps its left-to-right order, and every subject
  its side of center. \textbf{Bottom right:} gate clause~14 on
  the 32 cuts it can measure, counting the costume and prop-state values of
  what is staged on both sides of each cut. None disagree; every undeclared
  value is a prop's state, which the clause reports instead of passing.}
  \label{fig:transition-adherence}
\end{figure}

No pair of subjects reverses its left-to-right order at a cut. An audience does not notice three points of frame width but does notice a character who changes side of frame, the reversal the 180-degree rule prevents. Read off the staged mattes, across the 20 cuts with a subject on both sides every pair of subjects holds its left-to-right order, 32 of 32 pairs, and every subject stays on its own side of center, 44 of 44 crossings, including both at the two cuts in scene~11 where the lens turns, into the exchange and from shot to reverse shot (Figure~\ref{fig:transition-adherence}). Order survives every cut that moves or turns the lens, so this result is stronger than the drift result.

\subsubsection{Style continuity across cuts}
\label{sec:look-continuity}

After the style repair, no cut crosses the style break threshold. The failure for which a reader would reject a sequence lies in the generated image: adjacent panels that declare the same style and still do not look like the same hand. Measured by flat-region share, the 33 panels rendered before the style repair of Section~\ref{sec:visual-adherence} sat at a median 14.9\% flat and stepped a median 2.5 points across a cut; after it, the same 33 sit at 47.6\%, and all 42 current panels at 49.6\%, each within the declared style band. The absolute median step worsens to 5.2 points. Normalizing by the level of the two joined panels produces 14.4\% before and 9.0\% after, but this alternative denominator changes sharply with the intervention and is not plotted. At neither stage does a cut exceed 15 points. At scene~2's cut-in, however, the same seeds step 53\% of their level while the location description names a console the frame does not show, and 18\% without that one sentence. A seam that one sentence opens or closes belongs to the samples as much as to the style, and the threshold is chosen, not calibrated against a reader. These diagnostics therefore do not establish improved style \emph{continuity}, which stays open.

\subsection{A prompt-only baseline on an external corpus}
\label{sec:prose-baseline}

On an external corpus, greybox conditioning holds declared framing better
than either the director's own words or the compiled prompt. Four arms render the
same 204 shots at three seeds, for 612 renders per arm. Delivered head height
relative to the staged target is 1.906 from the director's own words, 1.733 from the
compiled prompt, and 0.955 with the greybox
(Table~\ref{tab:prose-baseline}). Seed-to-seed spread is respectively 0.057,
0.049, and 0.026, while declared size ordering survives in 82.0\%, 90.0\%, and
97.1\% of same-seed pairs. The compiled prompt is closer than the director's
own words on 336 of 522 paired cases (64.4\%, two-sided sign test
$p=5\times10^{-11}$), but both prompt-only arms exceed the target by more than
50\% in the median ratio. The greybox is closer than the director's own words on 530 of
555 pairs and closer than the compiled prompt on 499 of 507. The supported
conclusion is therefore that explicit geometry controls framing more reliably
than either wording, not that one wording is a satisfactory control.

The same corpus also carries the geometry stage's own checks, which need no generation. All 204 panels build, and the staged shot size equals the declared one on all 204, with the camera distance decided by height on 177 and by width on 27. The staged head rises monotonically through the seven declared sizes, from 0.064 of frame height at \texttt{establishing} to 0.776 at \texttt{extreme\_close\_up}, and every two-subject panel keeps its declared left-to-right order, 49 of 49. A single subject lands within 0.17\% of frame width of its declared position over 155 placements (mean 0.12\%), while the 98 placements in two-subject panels miss by a mean of 8.0\% and at worst 24.1\%: the same residual pattern under the fixed-staging solve, reproduced on a
corpus this paper did not author.

The 204 PROSE shots come from one production and contain one or two named
characters. The first comparison changes wording between the director's own words
and the compiled prompt. The second compares the prompt-only workflow (empty
latent, denoise 1.0) with the bundled greybox-conditioned workflow (flattened
greybox init, denoise 0.55) while retaining the compiled prompt. It therefore
tests the conditioned system as configured, but cannot separate the effects of
the control image and denoise setting. The
fourth arm declares a pose that the other three leave empty; it changes both the
staged body and one word of the prompt, so it is not a staging-only ablation, and
the resolver moves only 30 of the 204 shots, so its effect is measured on those 30
rather than on the corpus.
The English actions, character and location records, content classes, and 40
comparison labels were produced by a language model and were not human-reviewed. SAGE~\cite{sage2026} and
\pace{} ask different questions of the same released director storyboards; SAGE's system itself is not released.

Read the other way, the two axes trade against each other. The judge sees the described content in 87.1\% of the renders made from the director's own words, 67.0\% from the compiled prompt and 53.9\% from the greybox, and the ordering holds inside every content class: an action is seen in 96.2\%, 76.7\% and 57.5\%, a character state in 86.0\%, 68.0\% and 58.3\%, a scene state in 78.9\%, 57.8\% and 44.4\% (Table~\ref{tab:prose-baseline}). Paired on panel and seed the same order is significant in both comparisons, the director's own words over the compiled prompt on 133 pairs against 10 and over the greybox on 210 against 7. The arm that draws the most of what was described is therefore the one that holds the framing least, and the declared cast with it: the detector counts the declared number of people in 77.3\% of the director's-words renders, 93.7\% of the compiled ones and 99.2\% with the greybox.

The greybox arm's shortfall is concentrated where the description is an action, and the mechanism is visible in the build: \pace{} stages the actors through blocking and camera and leaves the action to the prompt, the first three arms declared no pose from the descriptions, so every proxy stood, and at a denoise of 0.55 the init held that standing body against the action the words asked for. That is a cost of staging a pose the description implies but the specification never declared, not a cost of staging as such. The judge has not been checked against human labels, so its rates are indicative and only the paired direction is read as a result.

A fourth arm declares the pose and tests that reading. Each row's action text is resolved to one of the staged proxies and written into the shot before the greybox is built, so the subject kneels or walks where the sentence says so, and the same 204 shots render at the same three seeds. The resolver moves 30 of those shots; the other 174 resolve to standing again, and they were re-staged and re-rendered all the same, so they are not held fixed but redrawn, and their pairs carry the sampler's variance rather than a treatment. Everything below is therefore read on the 30 shots the resolver moved, with the corpus-wide figure given beside it.

The framing is undisturbed. The delivered head comes back at 0.960 times its staged height against 0.955 without the pose, and the per-size ladder is unchanged to within six thousandths of frame height at every size. The seed-to-seed spread stays at 0.027 against 0.026 and the declared size ordering holds in 96.0\% of same-seed pairs against 97.1\%. Paired on the 30 treated shots the two arms are indistinguishable on how close the delivered head lands to its own staging, 47 against 35 ($p = 0.22$), which is the result this arm needs: the pose is not bought with the framing.

What moves is the content, and only where a body carries it. On the 30 treated shots the judge sees the described action in 74.4\% of the pose-staged renders against 58.9\%, and paired on panel and seed it prefers the pose-staged render on 16 of the 18 untied pairs ($p = 0.001$). Every one of those 16 is an action of a limb or contact; the character states and scene states among the treated shots tie on all nine pairs. The direction the mechanism predicts is the direction the measurement finds, and the effect is confined to exactly the class a staged body can carry. On those same 30 shots the compiled prompt alone draws the content in 76.7\% and the director's own words in 98.9\%, so declaring the pose carries the staged arm to within two points of the prompt-only arm while keeping the framing that arm loses. Across the whole corpus the same treatment reads 57.0\% against 53.9\%, 33 pairs against 14, because 174 of the 204 shots were given no pose to declare. The recovery is partial either way: 74.4\% is still below the 96.2\% the director's own words draw, and the reason is in the corpus, since 29 of the 30 shots the resolver moved are framed medium or tighter, where the body below the chest is out of frame and a declared pose has little surface to act on.

Two cautions travel with this arm. Declaring a pose changes both the geometry and one word of the compiled prompt, so it is a change to the specification rather than to the staging alone; and on the 174 shots that were redrawn without one, the same framing test comes out significant in the pose arm's favour (260 against 192, $p = 0.002$), which is a measure of what rebuilding and re-rendering alone can produce, and a reason to read this paper's other paired comparisons only where a treatment actually differs.

\begin{table}[tbp]
\centering
\footnotesize
\caption{The two axes separate in opposite directions, and declaring a pose moves the second while the first holds to within a point. Asked in the director's own words the described content is drawn most often and the declared framing and cast held least; with the staged greybox it is the reverse; the prompt \pace{} compiles from the typed fields sits between them on content and with the director's words on framing. 204 PROSE shots at seeds 101, 202 and 303, 612 renders per arm, one model. The two middle arms share one compiled prompt but differ jointly in sampler initialization and denoise: empty latent at 1.0 versus flattened greybox at 0.55. Their contrast therefore evaluates the configured greybox-conditioned workflow, not an isolated geometry or denoise effect. The right-hand arm adds a pose resolved from each row's action text, which changes the greybox and one word of that prompt. It moved 30 of the 204 shots, so the content rows above are diluted by the 174 it left standing, and the last row repeats them on the 30 alone: declaring the pose carries the staged arm most of the way to the compiled prompt's content without giving up any of its framing. Head heights are shares of frame height, medians over the renders with a head found; counts in parentheses are shots for a size and renders otherwise. Bold marks the best arm on a row; the judge is not checked against human labels, so its rows are read as directions rather than rates.}
\label{tab:prose-baseline}
\setlength{\tabcolsep}{4.5pt}
\begin{tabular}{@{}lrrrr@{}}
\toprule
Measure & Director's words & Compiled prompt & Prompt {+} greybox & {+} declared pose \\
\midrule
\multicolumn{5}{@{}l}{\emph{Framing, by detector against the staged head matte}} \\
Delivered $\div$ staged head height, median (IQR) & 1.91 (1.50--2.16) & 1.73 (1.55--2.07) & 0.955 (0.93--1.00) & 0.960 (0.93--1.01) \\
Head height, \texttt{establishing}, staged 0.064 & 0.092 & 0.271 & 0.101 & 0.101 \\
\quad \texttt{wide}, staged 0.106 & 0.147 & 0.344 & 0.130 & 0.127 \\
\quad \texttt{full}, staged 0.113 & 0.273 & 0.300 & 0.145 & 0.145 \\
\quad \texttt{medium}, staged 0.212 & 0.300 & 0.414 & 0.208 & 0.214 \\
\quad \texttt{medium\_close\_up}, staged 0.294 & 0.711 & 0.584 & 0.278 & 0.279 \\
\quad \texttt{close\_up}, staged 0.454 & 0.850 & 0.715 & 0.426 & 0.426 \\
\quad \texttt{extreme\_close\_up}, staged 0.776 & 0.832 & 0.698 & 0.730 & 0.730 \\
Declared size order kept, same-seed pairs & 82.0\% & 90.0\% & \textbf{97.1\%} & 96.0\% \\
Spread over three seeds, median CV & 5.7\% & 4.9\% & \textbf{2.6\%} & 2.7\% \\
Head found (612) & 591 & 541 & 576 & 578 \\
\multicolumn{5}{@{}l}{\emph{Cast, over the 603 renders outside crowd shots}} \\
People = declared cast, detector & 77.3\% & 93.7\% & \textbf{99.2\%} & 98.5\% \\
People = declared cast, VLM judge & 84.9\% & 97.8\% & \textbf{98.8\%} & \textbf{98.8\%} \\
\multicolumn{5}{@{}l}{\emph{Described content seen, by a VLM judge not checked against people}} \\
All descriptions (612) & \textbf{87.1\%} & 67.0\% & 53.9\% & 57.0\% \\
\quad an action: limb or contact (240) & \textbf{96.2\%} & 76.7\% & 57.5\% & 63.3\% \\
\quad a character state: expression or gaze (228) & \textbf{86.0\%} & 68.0\% & 58.3\% & 61.8\% \\
\quad a scene state: light or screen (90) & \textbf{78.9\%} & 57.8\% & 44.4\% & 40.0\% \\
\quad speech alone (54) & \textbf{64.8\%} & 35.2\% & 35.2\% & 37.0\% \\
\multicolumn{5}{@{}l}{\emph{The same, on the 30 shots the pose arm actually moved}} \\
All descriptions (90) & \textbf{98.9\%} & 76.7\% & 58.9\% & 74.4\% \\
\bottomrule
\end{tabular}
\end{table}

\subsection{Beyond the primary controls}

Seven further results bound what the primary controls may claim: entity consistency, the information content of the declared fields, breakdown fidelity, beat segmentation, a pre-render gate, visual adherence of the delivered panel, and an ablation of the identity plates.

\subsubsection{Entity consistency as a function of shot separation}
\label{sec:entity-consistency}

Scored against a different-entity floor, locations hold: same-location pairs score $+0.107$ on layout agreement and $+0.093$ on embedding similarity above the different-location floor (153 contrasting pairs). Character does not resolve: controlling for location change and gap collapses an apparent decay to $-0.004$ at gaps up to ten and $+0.032$ up to six. The whole-frame instrument detects no identity signal at this scale, which is not evidence that identity drifts.

\subsubsection{Information content of the declared fields}
\label{sec:field-provenance}

Several declared fields take one value across the whole corpus, so an adherence score computed against them measures self-consistency, not fidelity. Of 130 fields instantiated at least eight times (3267 instances), 38 take exactly one value and 23 more carry less than one bit of entropy; 14 are constant by nature in one film and 19 sit in groups this corpus does not exercise, and the finding is the remaining five. Screen position is the worked case: \texttt{screen\_\allowbreak position.\allowbreak y} is 0.52 on all 58 subject declarations, and \texttt{screen\_\allowbreak position.\allowbreak x} is array position on all 58, the breakdown assigning a zone by list index (\texttt{center}, then \texttt{center\_left}, then \texttt{center\_right}) and a compiler table mapping \texttt{center\_left}, \texttt{center} and \texttt{center\_right} onto 0.38, 0.50 and 0.62. The component that would compute a real position never runs, because \texttt{physical\_\allowbreak layout} is null in all 11 scenes, so a delivered position measured against 0.38 measures the geometry stage, not an intent. Three further cases failed one step earlier, and the corpus answers each: \texttt{setup.props[].state} takes four values across 54 of 87 prop declarations and reads \texttt{powered\_\allowbreak off} on every shot whose beat asserts a power loss, the 33 silent declarations confined to four props; \texttt{setup.environment.\allowbreak mood} was absent on 8 of 45 panels and is inferred; and the film's top matter and the director's shot design, which nothing read, are compared against the scene documents and report 32 divergences. What is missing is a per-field record of whether a value was derived from the screenplay, defaulted from a table, or authored by hand.

\subsubsection{Breakdown fidelity against the screenplay}
\label{sec:verified-breakdown}

What a language model asked to break a screenplay down in one pass leaves out cannot be seen from its output, and what it adds looks the same as what it read. The corpus's own breakdown was written that way, by a single call returning scenes, shots and starter panels, and measuring it against the screenplay is what this experiment does. The screenplay is parsed deterministically into 242 typed elements whose ranges all slice back out of the source; a model extracts entities, events, roles and state changes and must quote the words supporting each fact, and every quote is located by string search~\cite{gophercite2022,tanl2021,igcs2025}: 79 events, 56 entities and 19 entity states, with no ungrounded fact among the 154. The strictness has a price worth stating: 109 of the 154 quotes matched verbatim and carry real offsets, while 45, or 29.2\%, matched only after whitespace was collapsed, because the model reproduced the words and not the screenplay's line breaks. A located quote also proves less than it appears to, since it establishes that the span is in the screenplay and not that the span supports the value attached to it. Held to that ground truth, the segmenter places all 79 events into exactly one beat each, and all 22 state transitions survive segmentation.

An estimator from a different model family scores a breakdown against the same ground truth. The one-pass breakdown covers 0.37 of the events and omits 48, including the film's climax, the death of a principal character; a breakdown regenerated under the citation requirement covers 0.95, omits one, and cuts invented events from 13 to 4, with repeated runs differing by about $\pm 0.02$. Scoring the scorer is standard practice where no ground truth exists~\cite{fbi2024,mutationjudge2026}, and on 18 planted errors this estimator reaches a strict recall of 0.71 and an any-code recall of 0.93 over 14 measurable mutations, at a precision of 0.37. Eighteen is few: the rates are reported as the counts they come from rather than as estimates, since a proportion from 18 trials carries an interval tens of points wide, and comparable studies plant thousands~\cite{fbi2024}. The 0.95 is an upper bound, because a deleted action's event was re-attached to a neighbor instead of reported missing.

\subsubsection{Beat segmentation and state-checkable panels}
\label{sec:beats}

\begin{figure}[tbp]
  \centering
  \includegraphics[width=\textwidth,height=0.72\textheight,keepaspectratio]{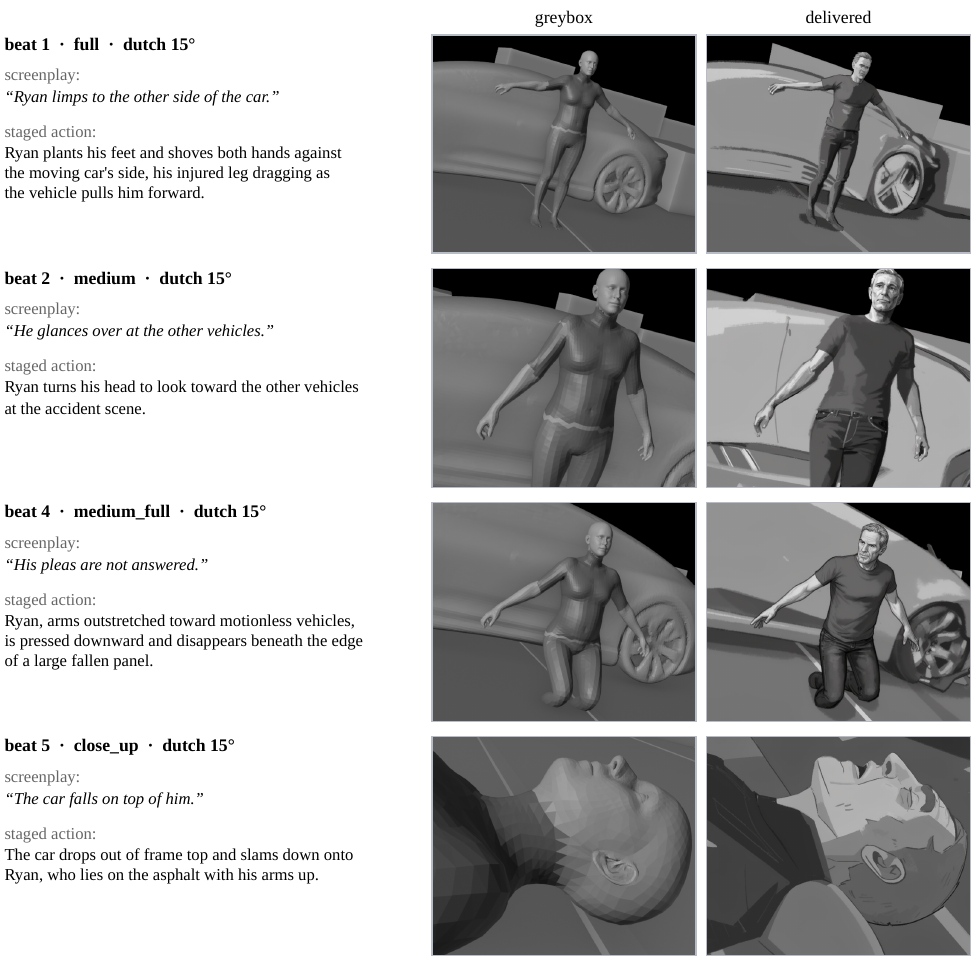}
  \caption{A curated human-in-the-loop pilot maps each beat of one scene to a
staged panel. It keeps the screenplay words, including the climax, which the
corpus's existing breakdown omits. Rows show the grounding words, staged action, control geometry and delivered image, cropped to the cast by one crop from the staged mattes. All rows share one seed and every camera is canted $15^\circ$: the dutch angle is the director's deliberate choice for the tension of this scene, declared in each shot's camera like any other value. Beat 1 is denoised at 0.45 instead of 0.55 so that it adds no unstaged bystander, and beat 5 is a close-up from above. These four panels are the paper's one exception to staging a body per character: they keep the shared SMPL-X proxy chosen by pose and stature, because their delivered frames were denoised from these greyboxes and pairing a restaged control with an old render would claim a correspondence it does not have. Nothing measured depends on it, since the solvers read the head matte and the projected read point and any proxy of the right height carries both. }
  \label{fig:beat-pilot}
\end{figure}

Segmenting at a change of dramatic state cuts the 79 events into 39 beats, since a panel should cover the smallest unit in which the world changes; each beat carries the state before it, its transition and the state after, read from a world-state timeline so that a later beat cannot quietly restore an irreversible state. The gain over asking a model to split the scene itself is not a better cut but a checkable one: a returned list of beats carries nothing to hold a panel to, whereas a beat that names the state before and after gives every panel built from it a claim the compiler has to discharge and the delivered frame can be read against. Structural boundary features are measured and only goal shift and reveal are judged by a model; where a weighted sum merged 40 of 41 adjudicated pairs, treating one strong signal as sufficient took the film from 26 beats to 39. Figure~\ref{fig:beat-pilot} is a curated human-in-the-loop diagnostic on
one scene, not an estimate of automatic success; its camera choices and two
panel descriptions were revised by hand: each beat compiles to a panel whose cast is the actors it shows and whose prop states come from its \texttt{state\_after}, and all six beats stage and render, the two without an actor by fitting the location's declared box. Edge agreement between greybox and delivered frame inside each staged silhouette runs 0.46 to 0.56, against 0.38 to 0.42 for the silhouette shifted away, a mean margin of $+0.10$ over three subjects; the check catches a relocated subject but not a body reversed end to end in place.

What the partition and persistence counts establish is well-formedness, not accuracy: a partition holds by construction, and any sound merge rule preserves the transitions, so neither count distinguishes this 39-beat cut from a 26-beat or a 79-beat one. Where the boundaries belong is not measured here. A segmentation is normally scored by WindowDiff and $P_k$ against a human reference~\cite{beeferman1999,pevzner2002}, with an inter-annotator ceiling beside it, and no such annotation exists for this corpus.

\subsubsection{Pre-render gate on the staged anchor}
\label{sec:greybox-gate}

Twenty-seven of the 42 staged panels pass the gate. Before rendering, it applies to the staged build what the anchor check applies too late. It has fourteen clauses. Seven check the build: every required subject has a matte, that matte clears an area floor, the cast keeps its declared left-to-right order, no subject hides another beyond a bound, the read point is visible, and no derived pass is older than the frame it belongs to. The seventh, whether the focal action reads, has nothing to check against while the subjects are static proxies with no rig, and is reported as unmeasured rather than counted as passed. Eleven panels fail subject occlusion, with a rear cabin passenger 35--50\% hidden behind the front row, and five composition clauses find five panels with a frame edge at a joint and none with touching outlines. One clause compares what a panel declares in frame with what is staged, and no corpus panel declares it yet. The last holds each panel against the one it cuts from inside its scene, on the costume each character wears and the state each prop is in: measurable at 32 cuts, failing at none, and reporting that 24 of them leave a value undeclared on at least one side. The gate reports at build time and does not refuse a render.

\subsubsection{Visual adherence of the delivered panel}
\label{sec:visual-adherence}

\begin{figure}[tbp]
  \centering
  \includegraphics[width=\textwidth]{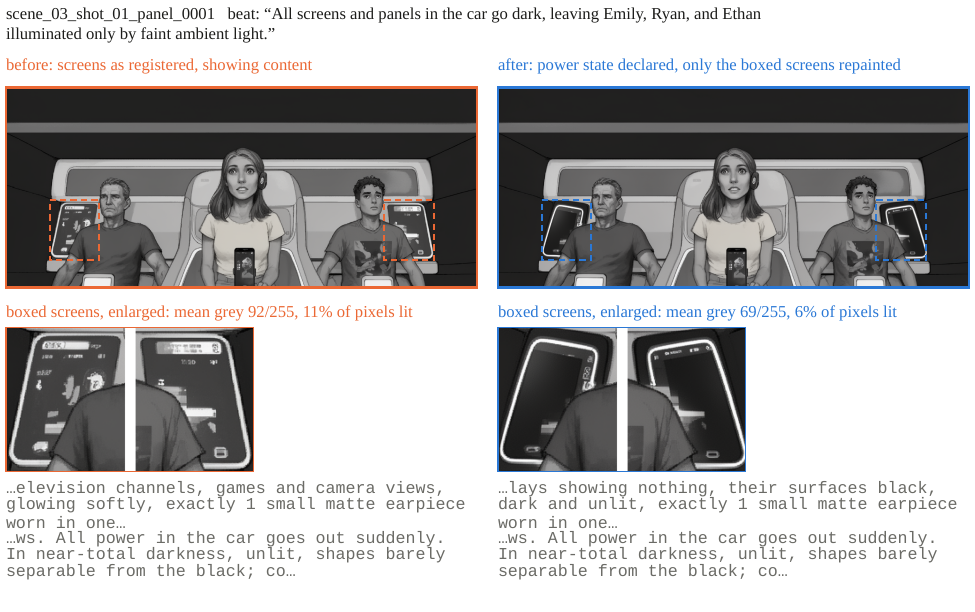}
  \caption{Declaring the power state blanks the screens that the registered appearance clause had showing content. The panel is the scene-3 shot whose beat declares a power loss, printed above it, shown as the three-person panel before the composition step split it into singles (Section~\ref{sec:shot-size-adherence}). \textbf{Left:} rendered with the screens' clause as registered; \textbf{right:} the same frame with only the boxed screen regions repainted at the same seed under the substituted clause, every pixel outside the boxes being the left frame's. The two prompts differ in that clause alone, and the lighting clause is the same on both sides, since a line-art style has no way to draw a light going out. Under each frame, the boxed screens are enlarged with their mean gray and the share of their pixels that are lit, above clauses excerpted from the prompt that produced it.}
  \label{fig:state-contradiction}
\end{figure}

Two failure modes appear on delivered pixels, and each has an intervention measured against it. First, a compiled prompt appended each prop's registered appearance clause to the beat unconditionally, so two clauses were composed as if both were always true: a beat asserting a transient state (a light going out, a console losing power) and a descriptor asserting the opposite permanent default (a panel that ``glows'', a cabin ``lit by sunlight''). A keyword scan of all 27 shots found two such contradictions, both rendered fully lit. The repair is a state channel: a per-shot lighting condition in which \texttt{blackout} overrides the natural source, and a per-prop transition, \emph{power: off}, that rewrites the appearance clause by substitution, since the diffusion model runs at a guidance scale of 1.0 and never evaluates a negative prompt. Figure~\ref{fig:state-contradiction} isolates the prop half at one seed; it shows that the channel can change what is drawn but gives no rate. A prompt is only half of it, since the sampler starts from the control image: a prop declared powered off is now toned dark in the greybox as well, which takes the declared-off surfaces of Figure~\ref{fig:one-scene}'s second panel from 112 in mean luminance, brighter than the rest of the frame, to 56 against a frame mean near 87, at the same seed, the same prompt and the same denoise. What the channel still does not govern is a display the diffusion model invents where the control image holds plain wall.

Second, every panel declares the same style field, but before the repair 23 of 33 panels came back as grayscale photorealism and ten as the requested line art; the greybox's flat-region share correlates with the delivered panel's at $r = 0.843$, so the control image's clay shading, not the style field, decided the style. The combined repair, flattening the control image and lowering the denoise
to 0.55, raised the median flat-region share from 14.9\% to 47.6\% over the same
33 panels. Because both variables change, this experiment establishes the
combined intervention but does not attribute the gain to either component.
Placement on the generated images, read by a pixel proxy, improved too: mean
error against the staged mattes fell from 9.4\% to 4.7\% of frame width and the
worst from 28.1\% to 15.3\%, with 25 of 33 improving. The keyword scan gives a
lower bound, and holistic human review of the delivered panels remains unrun.

\subsubsection{What the identity plates cost the staged composition}
\label{sec:refplate-ablation}

Conditioning the render on the cast's identity plates does not move the staging: over 42 panels rendered twice from the same greybox and the same seed, the delivered head comes back at 0.976 times its staged height without the plates and 0.968 with them, and 22 of the 42 land closer to the staged height with them against 20 further away (two-sided sign test, $p = 0.88$). The declared cast is recovered in 34 of 42 panels without the plates and 36 with them, three panels gained and one lost ($p = 0.63$).

The two signals compete for the same pixels, which is why the ablation is worth running: the cast is a declared field the renderer receives as a reference image alongside the greybox init, so pulling a face toward a registered likeness might pull the body away from where the greybox put it. At the level these instruments read, it does not. Both arms use the identical compiled prompt, the identical flattened greybox at denoise 0.55 and the panel's own corpus seed, and only the plates differ. What the plates buy is a different question: whether the delivered face carries the registered likeness is not measured, and no instrument in this paper reads identity.

\section{Conclusion and Future Work}
\label{sec:conclusion}

\pace{} contributes a typed path from screenplay evidence and authored
directing decisions to prompts, a metric scene, and artifact-level conformance
measurements. Its strongest guarantee is deliberately narrow: camera and
blocking fields discharged into geometry can be checked on the compiled camera
and staged render without a learned evaluator. Generated panels do not inherit
that guarantee. On an external corpus, greybox conditioning holds declared
framing substantially better than either the director's own words or a compiled prompt,
while reducing delivery of actions that the geometry does not stage. The result
supports explicit geometric control and also exposes its cost.

\textbf{Future work: a screenplay-to-storyboard model.} We designed the
specification for a use this paper does not make: as the annotation format for
a dataset built from finished films. A released film records the decisions this
corpus had to take from a model's action list and from cast order, namely where
each scene was cut into shots, each shot's size, angle, lens and camera
position, and where the actors stand relative to one another and to the set.
Aligning a screenplay's scenes to the film's shots and writing each shot as a
\pace{} record turns those decisions into training pairs. Existing film datasets
label part of this: MovieNet~\cite{movienet} and ShotBench~\cite{shotbench2025}
annotate shot type, and SkyScript-100M~\cite{skyscript2024} pairs scripts with
shooting scripts written in prose. A model trained on such pairs would output a
scene's shot decomposition, camera and blocking as records the compiler and
camera solver already consume, so a storyboard could be staged without planning
each shot by hand. It predicts records, not pixels, so its output stays
measurable by the audit of Section~\ref{sec:evaluation}; an annotation can be
checked the same way, by staging it and comparing the staged frame with the film
frame. This requires rights to the films, frame-level annotation of camera pose,
depth and people validated against professional shot lists and human
annotation, and a held-out comparison with a director's own decisions judged by
experts rather than by a model. The present experiments establish
none of it.

\end{document}